\PassOptionsToPackage{unicode}{hyperref}
\PassOptionsToPackage{hyphens}{url}
\documentclass[
]{article}
\usepackage[left=1in, right=1in, top=.75in, bottom=.75in]{geometry}

\usepackage[utf8]{inputenc} 
\usepackage{textgreek} 
\usepackage{amsmath,amssymb}
\usepackage{lmodern}
\usepackage{comment}
\usepackage{iftex}
\ifPDFTeX
  \usepackage[T1]{fontenc}
  \usepackage[utf8]{inputenc}
  \usepackage{textcomp} 
\else 
  \usepackage{unicode-math}
  \defaultfontfeatures{Scale=MatchLowercase}
  \defaultfontfeatures[\rmfamily]{Ligatures=TeX,Scale=1}
\fi
\IfFileExists{upquote.sty}{\usepackage{upquote}}{}
\IfFileExists{microtype.sty}{
  \usepackage[]{microtype}
  \UseMicrotypeSet[protrusion]{basicmath} 
}{}
\makeatletter
\@ifundefined{KOMAClassName}{
  \IfFileExists{parskip.sty}{%
    \usepackage{parskip}
  }{
    \setlength{\parindent}{0pt}
    \setlength{\parskip}{6pt plus 2pt minus 1pt}}
}{
  \KOMAoptions{parskip=half}}
\makeatother
\usepackage{xcolor}
\usepackage{graphicx}
\makeatletter
\def\maxwidth{\ifdim\Gin@nat@width>\linewidth\linewidth\else\Gin@nat@width\fi}
\def\maxheight{\ifdim\Gin@nat@height>\textheight\textheight\else\Gin@nat@height\fi}
\makeatother
\setkeys{Gin}{width=\maxwidth,height=\maxheight,keepaspectratio}
\makeatletter
\def\fps@figure{htbp}
\makeatother
\usepackage[normalem]{ulem}
\ifLuaTeX
  \usepackage{selnolig}  
\fi
\IfFileExists{bookmark.sty}{\usepackage{bookmark}}{\usepackage{hyperref}}
\IfFileExists{xurl.sty}{\usepackage{xurl}}{} 
\usepackage{url} 
\hypersetup{
  pdftitle={Ensoul: A framework for the creation of self organizing intelligent ultra low power systems (SOULS) through evolutionary enerstatic networks.},
  hidelinks,
  pdfcreator={LaTeX via pandoc}}

\title{\protect\hypertarget{_t2f9ru2h7duu}{}{}\textbf{Ensoul:} A
framework for the creation of self organizing intelligent ultra low
power systems (SOULS) through evolutionary enerstatic networks.}
\author{}
\date{}

\usepackage{titling}
\begin{document}
\maketitle
\vspace{-3em} 

\begin{center}
  Ty Roachford
  
  Florida Atlantic University
  
  Machine Perception and Cognitive Robotics (MPCR) Lab
  
  \href{mailto:Ty@TyfoodsForThought.com}{\uline{Ty@TyfoodsForThought.com}}
\end{center}

\hypertarget{section}{%
\section{}\label{section}}

\hypertarget{abstract}{%
\section{Abstract}\label{abstract}}

Ensoul is a framework proposed for the purpose of creating technologies
that create more technologies through the combined use of networks, and
nests, of energy homeostatic (enerstatic) loops and open-ended
evolutionary techniques. Generative technologies developed by such an
approach serve as both simple, yet insightful models of
thermodynamically driven complex systems and as powerful sources of
novel technologies. ``Self Organizing intelligent Ultra Low power
Systems'' (SOULS) is a term that well describes the technologies
produced by such a generative technology, as well as the generative
technology itself. The term is meant to capture the abstract nature of
such technologies as being independent of the substrate in which they
are embedded. In other words, SOULS can be biological, artificial or
hybrid in form.

\hypertarget{motivation-and-outline}{%
\section{Motivation and Outline}\label{motivation-and-outline}}

The contents of this paper are written with the intention of connecting
topics that are believed to help provide insight into fundamental
questions about life and intelligence as well as assist in motivating
the particular approach that this paper takes to creating generative
technologies. It is written at a high level and as such the precise
technical details of certain topics are left out. This has been done for
a number of reasons. The first is that this paper is meant to be rather
concise. Many of the precise details of each topic are not necessary to
understand the connection between topics. Secondly, it is meant to allow
for those without specialized backgrounds to more easily follow the
thread of discussion. This should make the paper as accessible as
possible. Readers who wish for a deeper understanding can refer to the
various references provided in this text.

The main questions that have motivated an evolutionary enerstatic
network approach to building generative technologies are the following:

\begin{itemize}
\item
  How do components of naturally evolved systems come together to work
  towards a common interest? In other words, how does self-organization
  work?
\item
  
  How do ``embedded agents'', that is, agents that are made out of the
  same ``stuff'' as the environment they are in, make decisions under
  logical uncertainty (i.e. with limited spatiotemporal access to the
  environment and limited processing power)?
  
\item
  
  How can one capture the complexity of nature in computer simulations
  without randomly searching the space of possible models, or solely
  using mathematical methods to create, tune and run highly
  parameterized biologically accurate models?
  
\end{itemize}

The major topics that this paper then synthesizes and proposes to
explore are:

\begin{itemize}
\item
  
  Energy homeostatic (enerstatic) loops as powerful models
  
\item
  
  The Free Energy Principle (FEP) as an informative description of such
  models
  
\item
  
  Open ended evolution through energetically closed systems of
  enerstatic loops
  
\item
  
  Assembly theory as an informative description of such closed systems.
  
\end{itemize}

Specifically, the paper is organized into eight sections as follows:

\begin{enumerate}
\def\labelenumi{\arabic{enumi}.}
\item
  
  The \hyperref[teleology-tame-and-souls]{\textbf{``Teleology, TAME and SOULS''}} section, details why it is
  helpful to talk about systems as having
  ``goals\textquotesingle\textquotesingle{} as motivated by the free
  energy principle, cybernetics and the Technological Approach to Mind
  Everywhere (TAME) framework. In that same context, the term ``SOULS''
  is then explained as being an ideal description of such
  substrate-independent phenomena.
  
\item
  
  The \hyperref[open-ended-evolution-and-assembly-theory]{\textbf{``Open-Ended Evolution and Assembly Theory''}} section
  describes what ``Open-Ended Evolution'' actually is, as it relates to
  one of the goals of the field of evolutionary programming. It further
  discusses open-ended evolution's relationship to assembly theory and
  posits a ``zoomed out'' look at assembly systems in general.
  
\item
  
  The \hyperref[ios-illusions-motivating-modeling-with-enerstatic-loops]{\textbf{''IOS Illusions: Motivating Modeling With Enerstatic
  Networks''}} section motivates enerstatic loops, networks and nests as
  an ideal abstraction for modeling SOULS as opposed to an ``Input
  Output'' system based approach.
  
\item
  
  The \hyperref[enerstatic-loops-networks-and-nests.]{\textbf{``Enerstatic Networks''}} section discusses how an
  enerstatic loop works with a simple code explanation, and also
  outlines how it can be specialized to model neural networks in
  particular whose generalization served as the birth of abstract
  enerstatic networks.
  
\item
  
  The \hyperref[evolving-enerstatic-networks]{\textbf{``Evolving Enerstatic Networks''}} section is the main
  contribution of this paper. It describes how open-ended evolution
  might be achieved through the use of energetically closed systems
  harboring enerstatic loops which naturally interact to form enerstatic
  networks and nests. As in the case of the ``Enerstatic Networks''
  section, as much detail as possible is given without prescribing
  particular implementation details.
  
\item
  
  \hyperref[thermodynamic-computing-enerstatic-networks]{\textbf{``Thermodynamic Computing \& Enerstatic Networks''}} explains
  what thermodynamic computing is, and posits that enerstatic loops,
  networks and nests may serve as an extensible general model system for
  their development.
  
\item
  
  \hyperref[taming-souls]{\textbf{``Taming SOULS''}} explains enerstatic networks as enacting
  cooperative and competitive dynamics whose consequences drive behavior
  and discusses approaches towards understanding and controlling such
  dynamics. In particular, the ``Test for Controlled Variable'' is
  discussed along with general outlines for a ``Test for Variable
  Controllability'' and a ``Test for Control Switch''.
  
\item
  
  \hyperref[discussion-research-directions-speculations-and-philosophy]{\textbf{``Discussion: Research Directions, Speculations and
  Philosophy''}} suggests possible research directions, speculates on the
  potential benefits of such research, and finally discusses the
  philosophical implications of deep scientific knowledge.
  
\end{enumerate}

All sections are laid out in a way that follows a more conversational
and ``story-like'' format, rather than ``just'' an academic style paper.
Papers with more technical details and specific implementation
methodologies along with empirical data from actual experiments are
underway. It was deemed important to get these connections out there
before the results from such experiments were completed in order to make
the ideas behind such work readily accessible to both the scientific
community, and the enthusiastic layman. The following is a small, but
exciting step into the future of technology, technology made with SOULS.

\hypertarget{teleology-tame-and-souls}{%
\section{Teleology, TAME and SOULS}\label{teleology-tame-and-souls}}

Any ''thing'' that exists for longer than 1 Planck length of time must
be ``doing something'' that allows it to persist its existence. This is
the most general form of the ``Free Energy Principle'' (FEP), which is a
theory that was initially proposed by Karl Friston to explain the
brain's behavior before being extended to living and non-living
``things''. In mathematical form, we say that a ``thing'' is a
``dynamical system'', specifically a dynamical system in a
``non-equilibrium steady state'' (NESS), and that what the system is
``doing'' allows it to keep itself in that state \textbf{\cite{10}}. The
``something'' that the system is ``doing'' can generally be called
``self-evidencing'', a term that Carl Gustav Hempel introduced with
respect to explanations for events. This is just what it sounds like,
the verb form of something being ``self-evident''. An explanation is
self-evidencing if the event which it explains provides key evidence for
the explanation \textbf{\cite{11, 30}}. It turns out that a big part of
the explanation for the persistent existence of every ``thing'' that we
can observe can be understood in terms of what it is doing. Simply put,
if a ``thing'' wasn\textquotesingle t engaging in this self-evidencing
process to at least some degree, then it wouldn't persist.

A related argument in the same vein, is that for something to exist --
despite not persisting it's existence -- it must necessarily have causal
power. In other words, it must be exerting some sort of effect on its
environment. This has been called the ``Eleatic Principle'', which was
first described in Plato's book ``Sophist'' \textbf{\cite{20}}. FEP
simply extends this to suggest that when looking at anything which is
able to \textbf{persist} its existence, then one would find, upon
measurement, that the causal power it is exerting must, on average, be
self-evidencing.

\begin{figure}[h!]
  \centering
  \includegraphics[width=5.61979in,height=2.65679in]{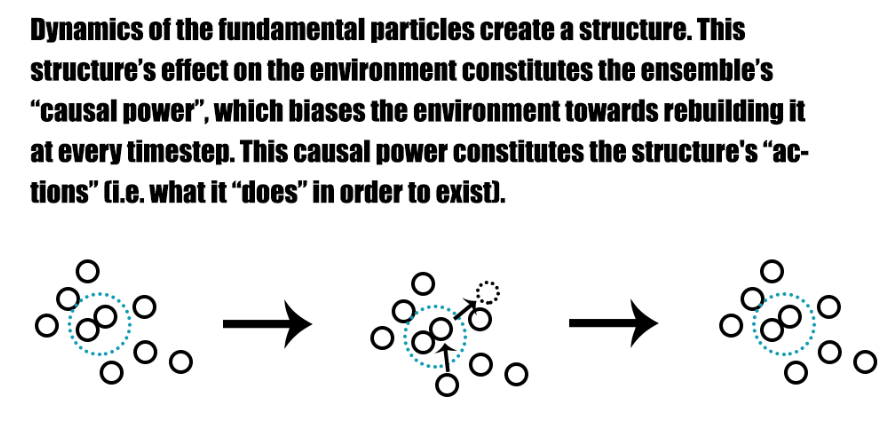}
\end{figure}

\begin{itemize}
\item
  
  Simplified illustration of ``Self-Evidencing Causal Power''. An
  environmental fluctuation results in ``too many'' particles inside the
  boundary, resulting in one of the bound particles being ejected.
  
\item
  
  A key phrase in the description is ``biases the environment''. It's
  not necessarily the case that the structure is capable of guaranteeing
  its own self-creation. It may be that the structure only increases the
  probability that the environment will re-create the structure for some
  duration of time.
  
\end{itemize}

By establishing a formal connection between the particular dynamics of a
``thing'' and variational inference, the FEP allows us to speak of
``things'' created via some self-organizing process, teleologically
\textbf{\cite{17}}. That is, to speak about a ``thing'' with respect to
its ``goal''. The term ``goal'' can be understood in a strictly
cybernetic sense. In a cybernetic system, the system's goal is to
regulate the value of a certain variable just as a thermostat regulates
the temperature. The system's regulatory behavior is considered to be
``goal-directed''. Since the most primary variable that a ``thing'' must
be regulating in order for it to persist its existence is its own
particular dynamics, we can indeed speak of its behavior as being
goal-directed. Thus, the FEP suggests that the primary goal of every
``thing'' is to self-evidence, regulate its particular dynamics, or most
specifically, minimize its variational free energy through the process
of variational inference. A detailed explanation on the mathematics
behind how the aforementioned formal connection is made is outside the
scope of this paper. Readers can refer to \textbf{\cite{17}} for more
information.

What is important about the construction of this formal connection
between the particular dynamics of a ``thing'' and variational inference
is that it serves as a guiding principle for better understanding the
behavior of any ``thing'' whose intelligence and cognition is related to
how well it can perform such self-evidencing variational inference. In
other words, all ``things'' exhibit some degree of intelligence and
cognition. Indeed, even the fundamental units, or particles, in any
theory of everything must necessarily have this ``self-evidencing
through inference'' characteristic in order to persist its existence
through time and thus has in the most minimal sense, a degree of
intelligence and cognition.

There are a few terms here that deserve further discussion, namely,
definitions of intelligence and cognition, as well as arguments for why
this paper prefers teleology over teleonomy. The definition of terms
intelligence and cognition have been fraught with arguments in a way
that is similar to what is meant by ``consciousness'' has. As such this
paper provides no strict definition of the two that everyone has agreed
upon. Instead, it takes an operational stance by defining them with
respect to the mathematics of variational inference underwriting the
FEP. When describing any ``thing'' the language of FEP speaks of its
``beliefs'', which essentially constitute knowledge that the ``thing''
has about the causes of its ``sensory perceptions''. It further suggests
that any ``thing'' must be ``doing something'', that is, taking
``actions'', upon its external environment, which are based on its
``beliefs''. In this paper, the transformation of ``sensory
perceptions'' and ``beliefs'' into ``actions'' upon an external
environment, which are on average in the service of self-evidencing, is
cognition. Cognition can then be measured via the degree to which that
``thing'' can ``sense'', ``believe'' and ``act''. In other words,
cognition can be measured via the answer to questions such as: ``What
does it sense and to what level of detail? What seem to be its
beliefs?'', and ``What are the actions available to it?'' respectively.
Intelligence, then, is a quality of this ``sense'', ``believe'', ``act''
cycle. In particular, intelligence is the adaptive capacity of a
``thing'' to accomplish its ``goal'' of existence. In other words, if we
see something that exists, then the closer that it can get to
non-existence (``death'') in some particular way (i.e. the more
``stressed'' it can get) and pull itself back into a ``healthy'' state,
the more intelligent we say that thing is. The ability of a ``thing'' to
pull itself back into a ``healthy state'' from a ``stressed state''
necessarily entails the ability of that ``thing'' to update its
``beliefs'' and therefore change its actions. This is because it must
necessarily do different things than that which resulted in that
``stressed state''. Thus, when that ``thing'' becomes better by some
measure at pulling itself back into a ``healthy state'' from a specific
type of ``stressed state'', then we will say that it has learned. The
various ways that a ``thing'' can be ``stressed'' indicate the various
ways in which a system has the opportunity to demonstrate intelligence,
thus intelligence is a multi-dimensional characteristic of cognition.

Teleonomy is a term that was developed to distinguish between ``things''
that have ``real goals'' (teleological systems), like humans, and
``things'' that only look ``as if'' they have ``real goals'' (telenomic
systems), like robots, or bacteria. This paper argues that this
distinction is unnecessary based on the formal connection that the FEP
made between a ``thing's'' particular dynamics and inference, the
application of the cybernetic sense of a ``goal'', and on the following
argument which focuses on the shared intrinsic characteristic of all
things, existence.

\textbf{Argument for fundamental teleology:}

\begin{enumerate}
\def\labelenumi{\arabic{enumi}.}
\item
  
  A ``goal'' is the desired result of a thing's behavior.
  
\item
  
  Identification of something's goal given its behavior is an unsolved,
  and in some cases potentially intractable, problem \textbf{\cite{23, 24, 25}}.
  
\item
  
  However, any ``thing'' that exists, must be exhibiting behavior which
  allows it to exist, in order to exist.
  
\item
  
  Thus, existence is a result of any ``thing'' that exists behavior.
  
\item
  
  The question is then: Does this ``thing'' ``intend'' to exist, is that
  its ``desired'' goal? -- We want to know whether this is the case or
  not, because we can only measure intelligence by observing behavior
  with respect to some goal. There are only two possible answers:

  \begin{enumerate}
  \def\labelenumii{\alph{enumii}.}
  \item
    
    Existence is its goal, in which case, observing its behavior allows
    us to measure its intelligence.
    
  \item
    
    Existence is NOT its goal, in which case, observing its behavior
    allows us to measure its \textbf{incompetence}
    
  \end{enumerate}
\end{enumerate}

While both answers to the ``intentional existence'' question posed in
point 5 are equivalent in what they allow for with respect to
measurement of a ``thing's'' behavior, the interpretation that each
allows is different. Only the first answer\textquotesingle s
interpretation properly aligns with what we commonly understand as
``intelligence''.

Readers may have noticed that there is a potential problem with this
definition of intelligence with respect to a ``thing's'' existential
goal. If you are observing some ``thing'' and it dies, does that mean it
was necessarily incompetent in some way? Would that further suggest that
``things'' which die voluntarily are incompetent? For example, what
about the case of a war hero who sacrifices himself to save his team?
The answer is that one must broaden the lens through which they are
observing. The definition of intelligence in this paper requires that
one properly identify the ``thing'' of study, and in the case of a
sacrificial war hero, such a sacrificial event cannot be properly
understood outside of the context that it is embedded in. Instead of
looking at the war hero as an individual ``thing'', this paper suggests
that, upon observation of such a sacrificial event, they must now be
looked at in context of the team, or collective, they are embedded in,
whose higher scale intelligence now becomes the ``thing'' of study. In
this context the war hero would not necessarily be considered
``incompetent'' because they are operating as part of a larger
organization, which may be the appropriate ``thing'' of study. An
analogy can also be drawn between the protective skin cells of an
organism, and the organism itself. The sacrificial behavior of skin
doesn't justify a judgment of incompetence. This sort of
existentially-based multi-scale analysis of intelligence has
consequences for what it means to be a particular individual ``thing'',
or ``self''.

You have a ``self'' when the components that make up any ``thing''
operate towards the same macro-level goal states \textbf{\cite{4, 6, 21}}. Most importantly, these components come to operate together only
through their natural physical behavior abiding by the path of least
action \textbf{\cite{16}}. An important part of establishing a coherent
``self'' is through the sharing of stress. The sharing of stress
enforces cooperation over competition by making defection an
impossibility. For example, in a situation where components are isolated
from each other in terms of stress, it is possible for component A to
stress component B without increasing its own stress levels. However, if
stress is shared between components via physical forces, or molecules,
then component A stressing any other component would increase that
component's own stress levels. The larger the degree to which stress is
shared between components the more of an impossibility that defection
becomes and the more coherent the ``self''. This binding of individual
``selves'' through stress can lead towards more resources, and therefore
computational power, being utilized towards the same aim \textbf{\cite{5, 7, 22}}. This might be summed up in the following phrase: ``When what
stresses you, stresses me, we are a self''.

As we can see from our previous discussion, the concepts of
``intelligence'', ``cognition'', ``goals'', ``stress'', and ``self'' are
all invariant across all ``things'' at every spatial scale from
fundamental particles to humans, and beyond. Additionally, these
concepts are completely naturalistic and operational in their
definitions. In other words, not only are they useful concepts through
which one can make measurements and perform engineering tasks, but they
also do not require anything that's not included in the laws of physics
at the most fundamental level. It is in this sense that the evolution of
the universe since its inception has been one in which these fundamental
properties have been scaled up through the combination of various
``things''.

The process through which more complex ``things" arise can be thought of
as a generalized form of evolution in which all ``things'', rather than
just ``living organisms'' specifically, are the object of natural
selection. In this generalized evolution, we speak about the stability
of ``things'' rather than the fitness of ``things''. Rather than saying
``Only the fittest survive'', we say ``Only the most stable persist''
\textbf{\cite{31}}. In other words, ``things'' that are more stable ``out
compete'' ``things'' that are less stable. This process of generalized
evolution is well described by assembly theory in what is then called an
``assembly process'' which will be discussed in more detail in the
``Open-Ended Evolution, Assembly Theory and Modeling'' section
\textbf{\cite{19}}.

Given that: (1) components of ``things'' come to operate together
through their own behavior (2) such ``things'' as well as their
components can be described as having ``intelligence'' (3) that the
behavior of all things are only following the path of least action, it
would seem that any ``thing'' that is assembled during an assembly
process can be best described as \textbf{self organizing} in that its
structure arises due to local interactions between its otherwise
independent parts, \textbf{intelligent} with respect to its adaptive
capacity to meet its existential goal, and \textbf{ultra low power} with
respect to how much energy it requires to meet that same existential
goal. This is why such structures are best described as Self Organizing
intelligent Ultra Low power Systems, or SOULS.

This paper focuses on what we, as humans, can do to assist in the
combination of ``things'' such that what we build results in practical
technologies that can be used for both science and engineering. As
previously seen, the invariant multi-scale characteristics intrinsic to
all ``things'' break down discrete categories by placing ``things'' on a
spectrum in which the experimenter, or engineer must ask ``How much''
and ``What kind'' of property does this ``thing'' that they are
studying, or engineering with, have \textbf{\cite{5, 6, 7}}? As such, the
terminology being used to refer to the ``thing'' in question best
reflects the goals of the observer rather than the intrinsic properties
of the ``thing'' itself. ``Things'' can be called ``structures'',
``materials'', or ``components'', in the case of engineering,
``chemicals'', ``organisms'', ``life'', in the case of biology, or, as
we will see in the case of Ensoul, enerstatic loops, networks and nests,
which are abstractions of SOULS at different scales.

Technological Approach to Mind Everywhere (TAME) is a framework being
developed by Dr. Michael Levin which aims to provide an
experimentally-grounded framework for understanding diverse bodies and
minds, or SOULS. The goal of the framework is to use such invariants as
a mode of thinking about how we can both create, and engineer with,
structures that have various levels of intelligence and cognition. An
important concept in TAME is the ``axis of persuadability''. Levin
states: ``Persuadability refers to the type of conceptual and practical
tools that are optimal to rationally modify a given system's behavior.''
Some SOULS are more persuadable than others, and their persuadability
is, in part, a function of what Levin describes as a SOULS' ``cognitive
light cone'' \textbf{\cite{5,6,7}}. The cognitive light cone represents
the spatiotemporal degree to which a SOULS' intelligence, or problem
solving ability, can take into account. For example, simple unicellular
organisms might ``care'' only about things like local sugar gradients,
and events that occur on the order of milliseconds to perhaps seconds.
In contrast, humans can care about what's going to happen to the
universe (huge space), in a few billion years (long time). Thus, the
cognitive light cone of a human is much larger than that of a simple
unicellular organism.

\begin{figure}[h!]
  \centering
  \includegraphics[width=5.53646in,height=3.30945in]{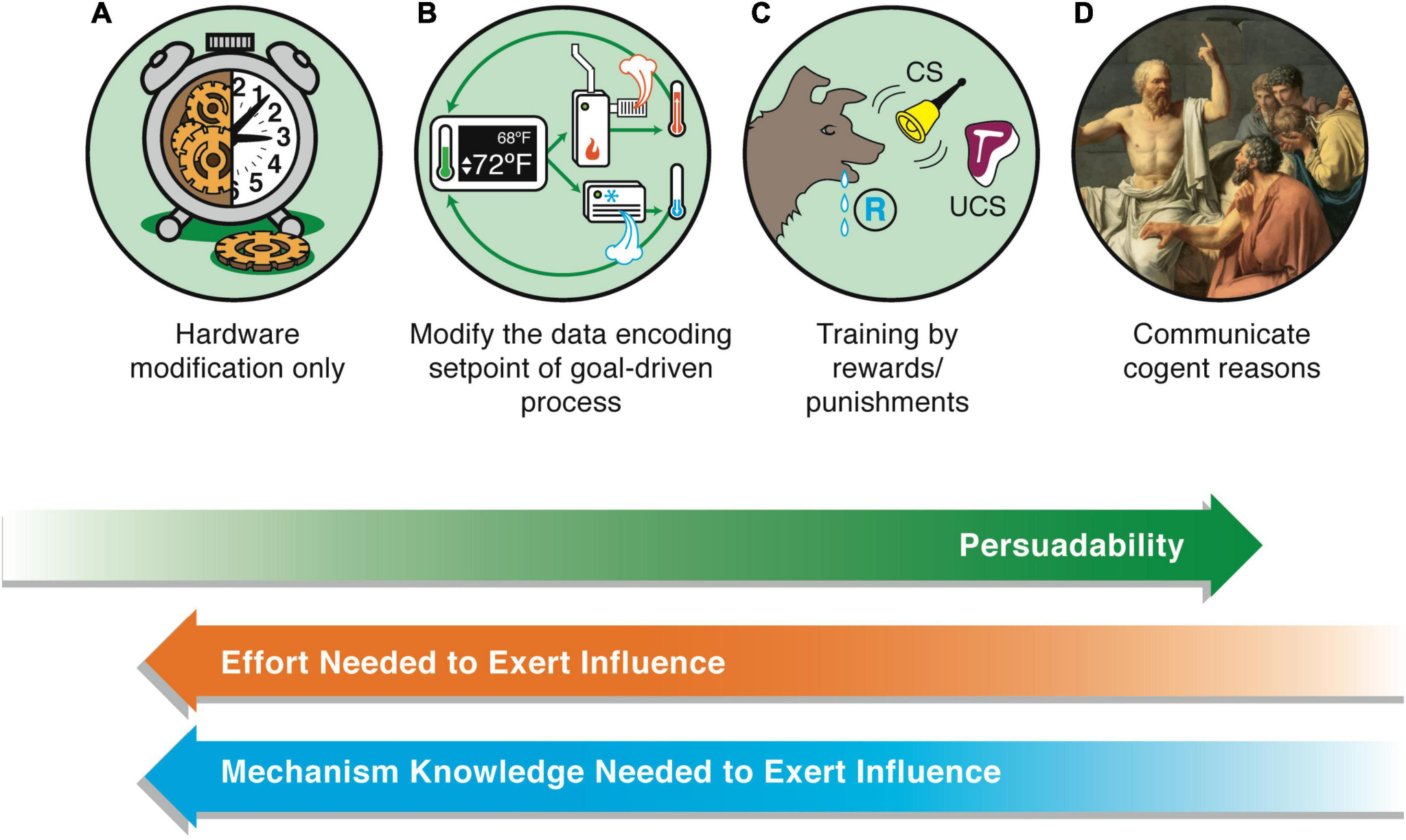}
\end{figure}

\begin{itemize}
\item
  
  "Figure adapted from Levin (2022), published in Frontiers in Systems
  Neuroscience, under the Creative Commons Attribution License (CC BY).
  Source:
  \href{https://www.frontiersin.org/articles/10.3389/fnsys.2022.768201/full}{\nolinkurl{https://www.frontiersin.org/articles/10.3389/fnsys.2022.768201/full}}"
  
\end{itemize}

{\centering
  \includegraphics[height=4in]{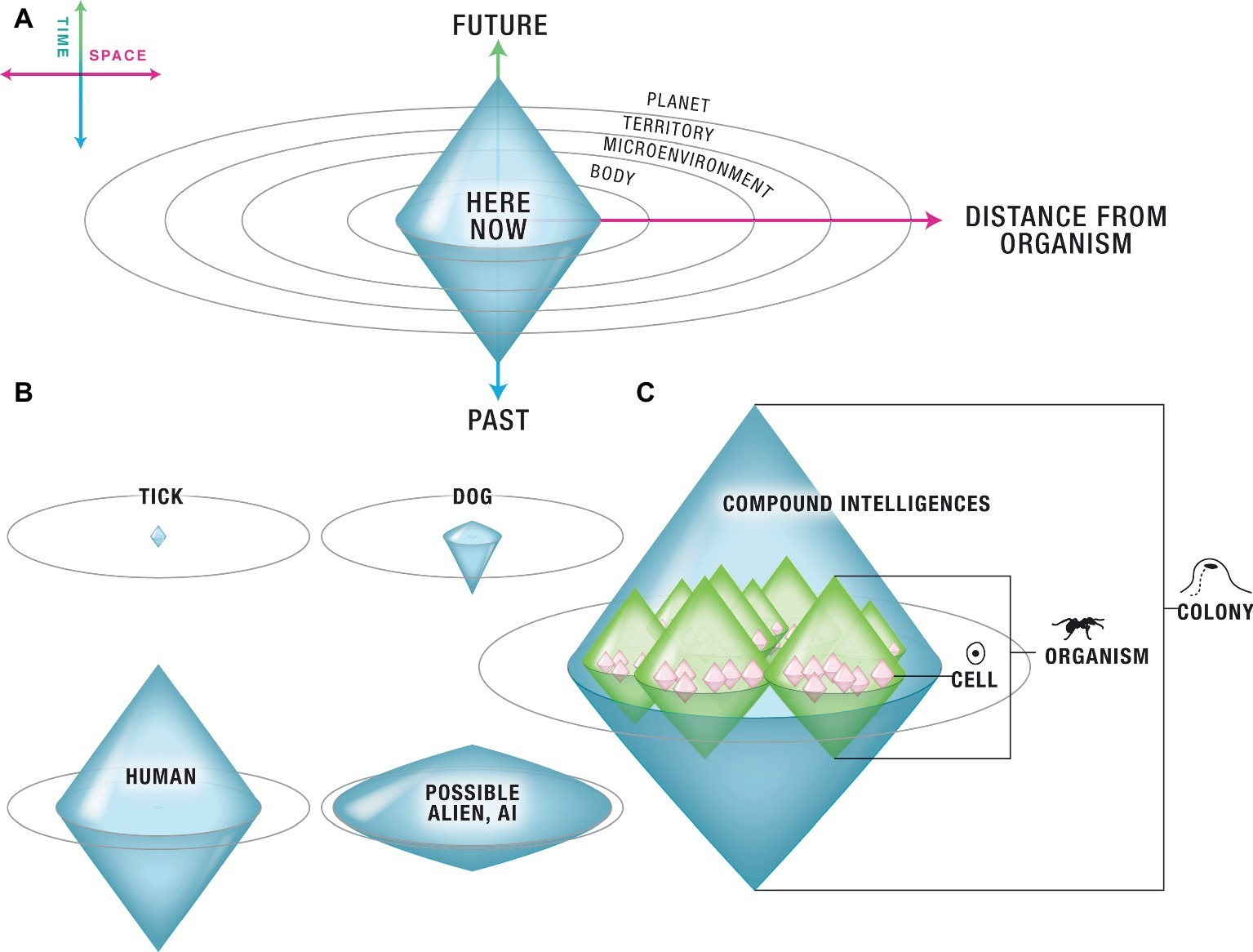}
\par}

\begin{itemize}
\item
  
  "Figure adapted from Levin (2019), published in Frontiers in
  Psychology, under the Creative Commons Attribution License (CC BY).
  Source:
  \href{https://www.frontiersin.org/articles/10.3389/fpsyg.2019.02688/full}{\nolinkurl{https://www.frontiersin.org/articles/10.3389/fpsyg.2019.02688/full}}"
  
\end{itemize}

For practical purposes, it is helpful to define the axis of
persuadability on a per-goal basis. Some SOULS are better than others
depending on the goal independent of their persuadability. For example,
if one's goal is for the SOULS to bake cookies, a typical computer would
need to be broken down into its constituent pieces and reconfigured
perhaps into a small cookie making robot. This would be low on the axis
of persuadability as it requires brute force micromanagement. However, a
working chef would only need to be told what to make and perhaps
convinced with some money. In this way, the axis of persuadability with
respect to some goal is anti-parallel with the amount of information
required to meet that goal. In other words, the amount of energy that
must necessarily be expended by the TAMEer in order to have the SOULS
meet some goal decreases as the SOULS' position on the axis of
persuadability moves to the right.

\begin{figure}[h!]
  \centering
  \includegraphics[width=6.5in,height=1.56944in]{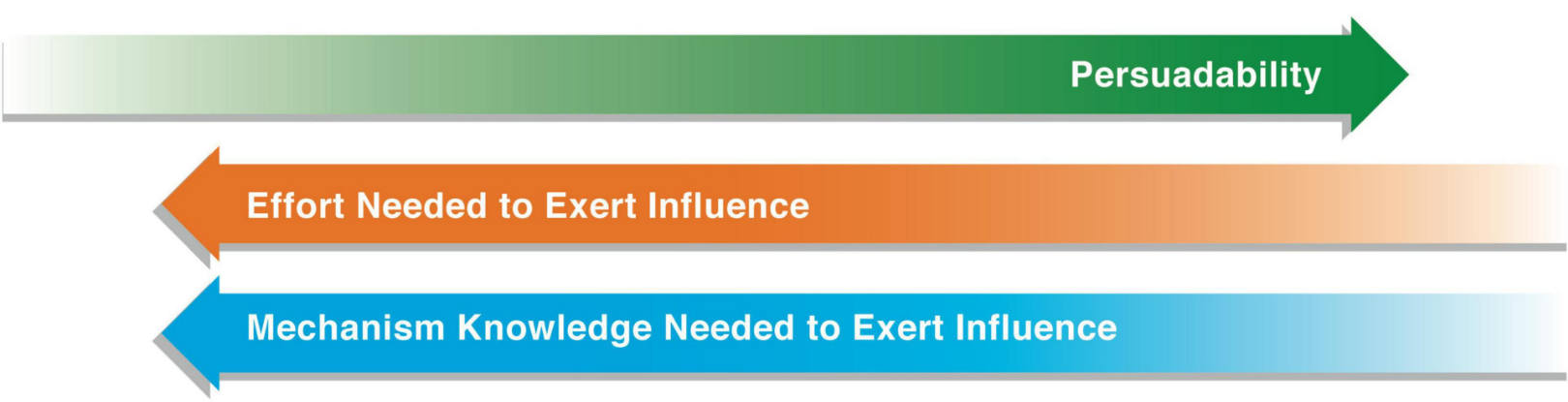}
\end{figure}

\begin{itemize}
\item
  
  Degree of persuadability vs the amount of energy, in the form of
  knowledge/effort, needed to get the SOULS to align with your goal.
  
\item
  
  "Figure adapted from Levin (2022), published in Frontiers in Systems
  Neuroscience, under the Creative Commons Attribution License (CC BY).
  Source:
  \href{https://www.frontiersin.org/articles/10.3389/fnsys.2022.768201/full}{\nolinkurl{https://www.frontiersin.org/articles/10.3389/fnsys.2022.768201/full}}"

\end{itemize}

It is important to recognize that the axis of persuadability is a
spectrum. We covered the extreme left, micromanagement, and the extreme
right, communication of cogent reasons, of the axis, but there are SOULS
in the middle as well. Two examples illustrated in the axis of
persuadability graphic are of a thermostat, and a dog. If one's goal is
to change the temperature of some room they are in, then the thermostat
is a good bet, and you don't need to micromanage all of its components
in order to do so, you only need to change the set-point. Similarly, if
one's goal is to have a dog bring you a beer from the fridge,
micromanaging all its muscles, or trying to control its neurons with a
brain computer interface, is probably not the best way to do it.
Instead, one can train the dog with treats (rewards). Intelligence is
truly in the eye of the beholder here. One must recognize how one's
goals align with the goal of the SOULS that they are interacting with.
The closer this alignment, the easier it is to ``convince'' the SOULS to
do what you'd like. More of TAME and other tools for identifying the
higher ``non-existential'' goals of a particular SOULS such that it can
be modified and worked with are discussed in the ``Taming SOULS''
section. We first turn to other concepts related to the development of
enerstatic networks as a tool for SOULS creation.

\hypertarget{open-ended-evolution-and-assembly-theory}{%
\section{Open-Ended Evolution and Assembly
Theory}\label{open-ended-evolution-and-assembly-theory}}

Despite the title, the field of evolutionary programming has not yet
been able to capture the true open-ended evolution that we observe when
we look around the universe. Dr. Kenneth Stanley suggests that the
reason is because most evolutionary algorithms are still focused on
optimization rather than divergence, and that the problem with
optimization is that the most ambitious goals are deceptive with respect
to their path to get there \textbf{\cite{8}}. In other words, if your
goal is to create something like Artificial General Intelligence (AGI),
then the path there is not at all straight-forward. In actuality, there
will be many deceptive looking sub-goals that appear to take you closer
to your goal, but actually take you farther from it. We see hints of
this when training a neural network to perform some task. The neural
network can move into a local minima with the ``belief'' that it's going
in the right direction, but of course, this is actually taking it
farther away from the global minima. A great practical example of
deception is the example of training a bipedal robot to walk. You can
either train it via an objective optimization approach where the goal is
to get closer and closer to walking, or you can do this with a divergent
explorative approach. Dr. Stanley employed both methods and found that
the biped ``trained'' via divergent exploration actually ended up
walking farther than the biped that was optimized towards walking
specifically \textbf{\cite{8}}. A more straight-forward example is that of
training a neural network to solve a maze. If you simply optimize for
minimizing the distance between the agent and goal, then the following
maze constitutes a deceptive example:

\begin{figure}[h!]
  \centering
  \includegraphics[height=3.25in]{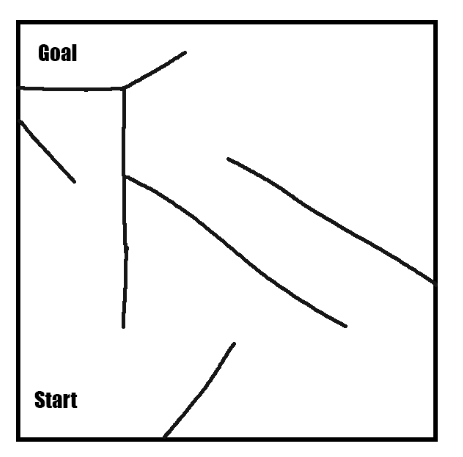}
\end{figure}

Most of technology has been focused on building machines, or programs,
that have particular desired outputs. That is to say, creation of
technology has mostly been focused on convergence, rather than
divergence. It is this focus on convergence that makes most of
technology today ``evidence of life'' rather than ``life'' itself. In
assembly theory, ``evidence of life'' refers to structures with an
``assembly index'' that is higher than a certain threshold. The assembly
index is an intrinsic measure of a structure's complexity
\textbf{\cite{1, 2, 3}}. The idea is that high complexity structures have a
dramatically smaller chance of being created through random unbiased
processes, than through biased processes and thus constitute evidence of
some stable constructor. That stable constructor would be considered
``evidence of life'' and indeed a living structure, or alternatively, a
SOULS lying relatively far to the right on the cognitive continuum. The
more complex the structure the SOULS is capable of constructing, the
more ``alive'' that SOULS could be said to be. Thus, placing the notion
of ``alive'' on a continuum.

Zooming out, in the context of assembly theory, life itself could be
described as the process by which evidence of life, at potentially
varying levels of complexity, is generated. In some sense then, our
universe is ``alive''. Since assembly theory is a physical theory, we
can think about other possible universes that may be more or less
``alive'' than our universe. One could say that the level of
``aliveness'', ``interestingness'', or ``intelligence'' of a universe
would depend on that universe\textquotesingle s trajectory through
``assembly space''. Specifically, the most alive, interesting and
intelligent universe would be one whose trajectory, on average, best
maximizes the number of unique objects at each possible level of
complexity. Interestingly, as a consequence of the FEPs description of
the informational symmetry of physical interactions, the environment
(i.e. the universe) of any structure must itself be considered an agent
(which we will call ``U''). Furthermore, as U gains copies of the
various structures within it, it is better able to predict the behavior
of those structures, and therefore it is better able to minimize its
variational free energy (VFE) \textbf{\cite{18}}. If one does indeed
grant that intelligence is the capacity of an agent to minimize its VFE,
thereby persisting its existence, then the most intelligent universe is
one in which the number of copies of each existing unique object is
maximized. A more rigorous treatment of measuring aforementioned
qualities in assembly systems is outside the scope of this paper.

{
  \centering
  \includegraphics[height=3.25in]{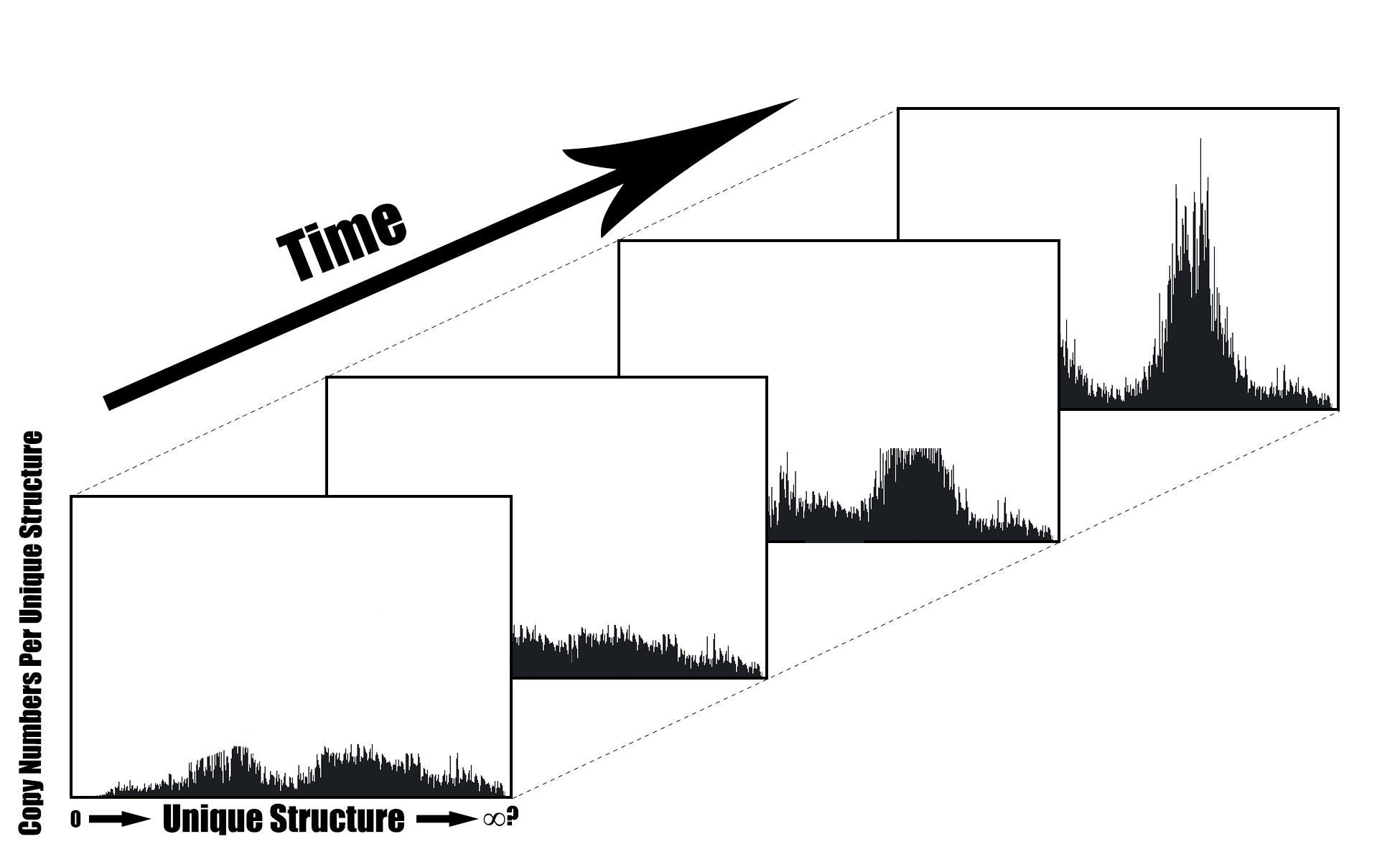}
  \par
}

\begin{itemize}
\item
  
  Demonstrating what the trajectory of an intelligent universe through
  assembly space might look like. On the x-axis, we have indexes
  associated with each unique structure. On the y-axis, we have copy
  numbers associated with each unique structure I.E. how many of that
  structure is currently present. Each frame represents a different
  point in time. Each structure-index pair is ordered such that
  structures towards the right have a greater complexity than structures
  on the left. This forms a sort of ``wave'' where structures bootstrap
  the creation of more complex structures, which additionally represents
  resource constraints. In other words, under resource constraints, more
  complex structures preclude the existence of a high number of less
  complex structures (hence the wave). However, without resource
  constraints, this wave could in principle propagate forever; Hence the
  infinity symbol-question mark pair on the bottom right side of the
  graph. It should further be noted that such infinite propagation would
  not necessitate a decrease in less complex structures thus increasing
  perhaps, the ``interestingness'' and ``aliveness'' of the system;
  Interestingness being the diversity and amount of existing structures
  in general, and aliveness being the diversity and number of living
  structures, or stable constructors, specifically.
  
\end{itemize}

Life itself is thus the process through which SOULS assemble and
recombine to create more SOULS of varying complexity (evidence of life
that might itself be living). The suggestion here is that we should
focus more on how to create technologies that create more technology. In
other words, how to create intelligent, open-ended, evolutionary, life
processes. This will help us avoid deceptive sub-goals, discover new
technologies, and through the study of structures created by such a
process, learn more about existing structures, or SOULS.

An interesting point to note here is that an assembly system itself
exhibits \textbf{self-organization} with respect to consequences of its
internal dynamics, \textbf{intelligence} with respect to its ability to
create and sustain high complexity objects and \textbf{ultra low power}
in that the ``running'' of its dynamics (i.e. execution of causal power
by its components) require a total of 0 energy \textbf{\cite{53}}. Thus,
an assembly system itself can be considered a Self Organizing
intelligent Ultra Low Power System, or SOULS.

The requirements for open-ended evolution have been strongly hinted at
through biological experimentation and observation \textbf{\cite{6, 8}},
and are well described phenomenologically by assembly theory
\textbf{\cite{2}}. Evolution doesn't just employ random mutation, but
biased mutation through mechanisms including, but not limited to,
enhanced epigenetic capacity, transposition, horizontal gene transfer,
and hybridization. Such mechanisms allow organisms produced by evolution
to essentially participate in their own evolution through increasingly
intelligent mutation operations. This reciprocal biased process is well
described by assembly theory.

According to assembly theory, there is information to be gained about a
so-called ``Assembly Space'' by considering the discovery process of
assembling novel objects, i.e. the rate at which new objects are
discovered (k\textsubscript{d}) over some time (discover time scale -
t\textsubscript{d}), and the rate of production of a specific object
(k\textsubscript{p}) over some time (production time scale -
t\textsubscript{p}). When the ratio
k\textsubscript{d}/k\textsubscript{p} is much greater than one, this
results in an explosion of unique objects with a low number of copies
(low copy number), whereas when k\textsubscript{d}/k\textsubscript{p} is
much higher than one, we observe a high abundance of some common objects
with high copy number. Either extreme leads to the absence of
trajectories that grow more complex objects \textbf{\cite{2}}. It can be
demonstrated that the appearance of open-ended evolution in a physical
system can be detected by a transition from k\textsubscript{p}
\textless{} k\textsubscript{d} to k\textsubscript{d} \textless{}
k\textsubscript{p}, in other words from no-bias towards what gets
created next, to increasingly more bias towards what gets created next.

\begin{center}
  \includegraphics[height=4in]{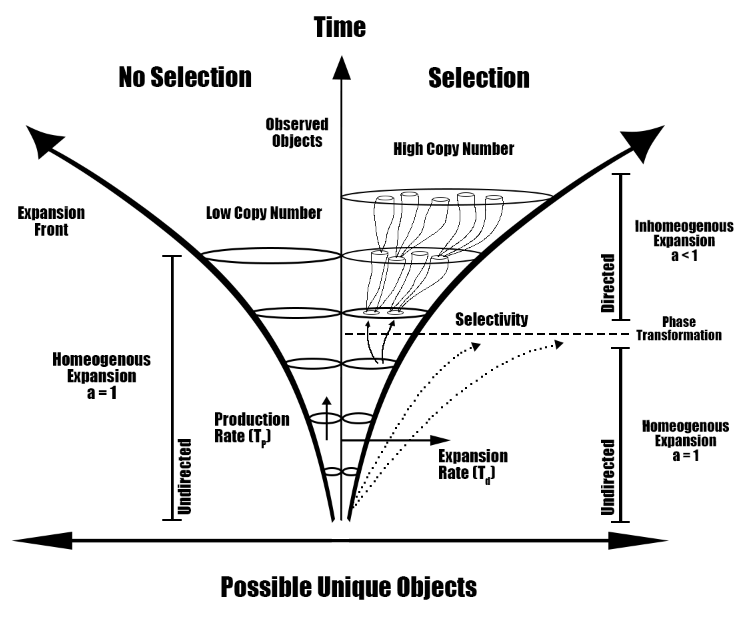}
\end{center}

\begin{itemize}
\item
  
  Graphic demonstrating the transition from unbiased to biased
  complexity growth that occurs under selection. Figure has been
  recreated from ``Assembly Theory Explains and Quantifies the Emergence
  of Selection and Evolution'', Source:
  \href{https://arxiv.org/ftp/arxiv/papers/2206/2206.02279.pdf}{\uline{https://arxiv.org/ftp/arxiv/papers/2206/2206.02279.pdf}}
  \textbf{\cite{2}}
  
\end{itemize}

Essentially, assembly theory asserts that open-ended evolution
algorithms follow the format: unbiased exploration to generate unique
low-level building blocks which potentially limit the possibility of
other low-level building blocks, followed by the combination of those
low-level building blocks into ``higher level building blocks'', some of
which are capable of sustaining themselves across time through their
intrinsic causal dynamics. The two questions that this raises are:

\begin{enumerate}
\def\labelenumi{\arabic{enumi}.}
\item
  
  What are those low level-building blocks?
  
\item
  
  What is responsible for the combination, and selection, of those
  building blocks into higher level building blocks?
  
\end{enumerate}

A helpful analogy is to see low-level building blocks as analogous to
primitives in a programming language, and to see the entity responsible
for combination, and selection, of those building blocks as analogous to
software engineers constrained by the syntax, or rules, of the
programming language. Software engineers combine these primitives into
ever more complicated functions, which are then selected based on their
usefulness and used to make ever more complicated pieces of software.
Importantly, the programming language being used evolves and diversifies
as do the ideas and models through which the software engineers operate.
In actuality, this ``analogy'' is rather just a subset of a larger
assembly process that is the evolution of the universe. The focus of
this paper is how we can model an assembly process that captures the
evolution of the universe ``as it could be'' using enerstatic loops. But
first, we start by motivating why such loops are an ideal candidate for
utilization as our low-level building blocks, explain how they work and
show an example use case.

\hypertarget{ios-illusions-motivating-modeling-with-enerstatic-loops}{%
\section{IOS Illusions: Motivating Modeling With Enerstatic
Loops}\label{ios-illusions-motivating-modeling-with-enerstatic-loops}}

A consequence of the fact that all ``things'' that exist must be ``doing
something'' in order to exist (I.E. self-evidencing) is that most
``things'' are not best described as ``Input/Output Systems (IOS)'' like
most typical computers are. Instead, they are best described as SOULS
that are regulating their own particular dynamics and in doing so, exert
causal power on the environment which appears to be its output in
response to some input.

The difference between IOS and SOULS is best illustrated by how each
system solves the problems that are presented to them. IOS solve their
problems using analytical solutions whereas SOULS solve their problems
using non-analytical solutions. Analytical solutions to any given
problem constitute modeling that problem explicitly where each
``significant'' part of the problem has a corresponding part in the
model. In this case, the ``thing'' in question has an explicit model of
the problem to be solved. Non-analytical solutions constitute lawful
correspondence between a ``thing'' and its problem such that the natural
dynamics of the parts that make up that ``thing'' solve the problem in
question without explicit representation of that problem. In this case,
the ``thing'' *is* a model of the problem to be solved. These two
problem solving methods have also been referred to as ``weak
anticipation'', and ``strong anticipation'', respectively
\textbf{\cite{33}}.

A great example of these two very different problem solving methods is
their application on the ``outfielder problem'', which asks: ``How do
outfielders catch a fly ball?''

\begin{figure}[h!]
  \centering
  \includegraphics[width=6.5in,height=2.77778in]{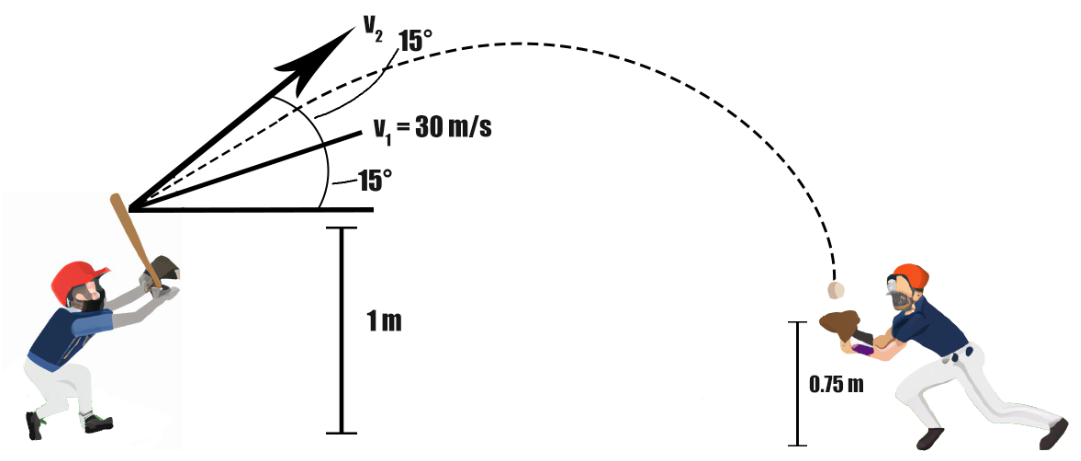}
\end{figure}

\begin{itemize}
\item
  
  Illustration of the ``Outfielder problem'' using an ``analytical''
  solution where one measures the initial conditions of the baseball and
  calculates its trajectory using equations for projectile motion.
  
\end{itemize}

An analytical solution to this problem might constitute the creation of
a mathematical model using representative equations. For example, a
``thing'' could estimate the force and angle at which the ball was hit
by the baseball player, and plug those values into equations that
represent projectile motion. In this scenario, the ``thing'' must
represent each input variable for the aforementioned equation, as well
as the equation itself, with corresponding physical parts. The dynamics
of these parts would constitute the calculation of the output which
would then have to be used to drive behavior.

A non-analytical approach to the outfielder problem would be to choose
some particular aspect of the projectile motion of the baseball and
control that variable such that its measured value remains fixed in such
a way that leads to one's desired result. For example, a ``thing'' could
look at the ball flying through the air and attempt to keep the speed of
the ball constant in their visual field such that the ball does not
appear to speed up or slow down. The idea is that if the ball is
accelerating, then it will land behind the ``thing'', and if the ball is
decelerating then it will land in front of the thing. In this scenario,
the ``thing'' doesn't require explicit representations; the ``thing''
only needs to be able to measure changes in speed, and use such changes
as feedback for driving behaviors that result in decreased measured
change. This proposed solution is called the ``Optical Acceleration
Cancellation'' approach and was first put forward by Seville Chapman
\textbf{\cite{34}}.

The distinction between an analytical and non-analytical approach is
crucial when it comes to understanding and modeling the ``thing'' in
question. It shifts the focus from ``What algorithms is this thing
implementing?'' to ``What variables is this thing controlling?''.
Indeed, treating a SOULS as if it is an IOS is a categorical error that
can lead to frustration because it can lead one to believe that the
behavior of the ``thing'' is highly variable either due to noise present
in its input, or intrinsic non-determinism \textbf{\cite{16}}. The
benefits that natural systems receive for engaging in non-analytical
methods over analytical methods are clear. Non-analytical methods are
simpler, faster, more adaptive and less power hungry while still
benefiting from much of the power that prediction has to offer
\textbf{\cite{33}}. They accomplish this by avoiding explicit
representation of parts of the problem being solved and by avoiding
using energy to calculate, and re-calculate, speculative analytical
solutions. This suggests that SOULS, rather than IOS, are likely to be
the more prevalent solution in nature \textbf{\cite{48}}.

Indeed, although IOS do exist in both nature \textbf{\cite{34}} and
technology, the parts that make up those IOS consist of a collection of
SOULS whose ``highly variable behavior'' are squelched as much as
possible so that they can in effect behave as an IOS. In computer
engineering, this ``highly variable behavior'' goes under the name of
``noise'', and comes in many different flavors such as thermal noise,
flicker noise, shot noise and quantum noise. Combating such noise is an
important part of building fast, reliable computers and thus techniques
such as filtering, shielding, grounding, the addition of noise margins,
error-correcting codes and clock synchronization are all used towards
that end. Indeed, combatting noise due to thermal fluctuations is so
important that it has been cited as the leading candidate for the
ceasing of the exponentially accelerating rate of technological progress
known as Moore's law \textbf{\cite{36, 37}}.

This paper suggests that ``highly variable behavior'', or ``noise'', in
a ``thing'', whether that ``thing'' is a biological organism, or a
circuit in a computer, is the observed result of a categorical error. In
other words, it is the result of viewing a ``thing'' as an IOS, when it
is better viewed as a SOULS. In this sense, IOS are illusions that are
approximated to arbitrary degrees of reliability by SOULS in the same
way that the ``actual value'' of some irrational number like \textpi\ can only
be approximated to arbitrary degrees of accuracy. This has been referred
to as the ``behavioral illusion'' under ``Perceptual Control Theory''
(PCT) \textbf{\cite{38, 39, 40}}. Does this mean that we should abandon the
idea of an IOS-based approach in favor of a SOULS-based approach? This
paper suggests otherwise. Instead, just as we appreciate the idea that
concepts like ``intelligence'', ``cognition'', ``goals'', ``stress'',
``self'' and ``life'' lie on a continuum, so should we appreciate the
spectrum between idealized SOULS and idealized IOS. We can refer to the
degree to which something modifies its own input simply as ``feedback'',
with zero feedback representing an idealized IOS.

Here, ``idealized SOULS'' refers to a SOULS that perfectly regulates the
variable it is controlling, such that the controlled variable simply
cannot be perturbed because any possible perturbation has already been
``predicted''. This is akin to mathematical models of anticipatory
synchronization in which systems A and B are perfectly coupled such that
A always predicts B no matter its behavior \textbf{\cite{33}}.
``Idealized IOS'' refers to an IOS that is completely deterministic and
perfectly reliable. The reference to ``idealized'' is an acknowledgement
that there does seem to be noise in most systems whether they are IOS,
or SOULS. For example, in the middle of this spectrum, we might imagine
an odd thermostat. This thermostat controls the temperature of the space
it is embedded in 50\% of the time, while the other 50\% of the time it
forwards its computations to some other system that does who knows what.

This perspective suggests that when it comes to trying to understand any
``thing'', we should first assess whether that ``thing'' is very much
like a reliable IOS, or more like a SOULS. As with assessing the level
of any ``thing's'' intelligence, the scale of analysis at which this
assessment is made, in addition to the careful partitioning of our
observed sensory data into ``things'', is critical to a proper
understanding. A too detailed look at an IOS, no matter how reliable it
is, will reveal SOULS-like behavior, in other words, one would observe a
collection of ``things'' that are regulating their own dynamics -- A
failure of appropriate choice of scale. Similarly, paying attention only
to particular pieces of a SOULS can result in the conclusion of simply a
collection of noisy IOS -- A failure of appropriate partitioning.

Given that SOULS, and not IOS, seem to capture things at the most
fundamental level, and that IOS can be approximated arbitrarily well by
a collection of SOULS, this paper suggests the usage of a simple
abstraction of SOULS in order to model any ``thing''. This simple
abstraction is an energy homeostatic loop, or in other words, a
cybernetic control system that is chiefly concerned with managing its
own ``energy level''. The big questions that explorations with this
abstraction hopes to provide insight to are things like:

\begin{enumerate}
\def\labelenumi{\arabic{enumi}.}
\item
  
  If everything is made up of one ``thing'' (energy), how and why do we
  come to observe distinct things?
  
\item
  
  How might a differing method of the partitioning of energy into
  ``things'' be helpful?
  
\item
  
  How do collections of energy homeostatic, or enerstatic loops, come to
  regulate variables at higher scales (i.e. measurable variables at
  larger spatial scales like the speed of a ball)?
  
\item
  
  To what degree is it possible to control and get along with such
  collectives?
  
\end{enumerate}

\hypertarget{enerstatic-loops-networks-and-nests.}{%
\section{Enerstatic Loops, Networks and
Nests.}\label{enerstatic-loops-networks-and-nests.}}

The Ensoul approach is agnostic as to what form an enerstatic loop can
take. The key is only to create a negative feedback system, or control
system, that regulates its own ``energy level'' in such a way that
exerts causal power on its environment. In other words, to create an
enerstatic loop. The energy level is a partial representation of the
state of the enerstatic loop, while the causal power is a representation
of the particular energy exchange between the enerstatic loop and its
environment, thus making such a loop a ``thermodynamically open
system''. We are not concerned with how enerstatic loops come to be,
instead we make the observation that ``nature builds where energy
flows'' and build our models accordingly. It is critical to not mistake
the following specific example implementations as the final, and only
way, of creating enerstatic loops. One can think of these only as
thought experiments within the Ensoul framework that happen to be
implementable as code on a computer.

An enerstatic loop can connect to other enerstatic loops via energy
exchange which together constitute an enerstatic network. Those same
networks, or singular loops, can be present inside of larger enerstatic
loops, which constitute enerstatic nests. This captures the multi-scale
fractal nature of reality, however, it can make the terminology
confusing. Given such a multi-scale approach, we distinguish between the
``main'' enerstatic loop that is the subject of study/modeling, and the
``lower level enerstatic loops'' that are nested inside of our main
loop, by calling such nested loops ``structures''. We model with
enerstatic loops, networks and nests using the following guidelines:

\begin{enumerate}
\def\labelenumi{\arabic{enumi}.}
\item
  
  There is a flow of energy into the system that depends on the action
  of that system.
  
\item
  
  This energy flow results in the establishment of an energy
  distribution, where energy is higher and lower in different parts of
  the system in the form of other enerstatic loops, or ``structures''.
  The establishment of this energy distribution represents the system
  starting out ``alive'' I.E. in a non-equilibrium steady state.
  
\item
  
  This energy flow results in the aforementioned energy distribution
  being modified in some way.
  
\item
  
  There is an inevitable loss of energy as the system generates entropy
  via exertion of causal power.
  
\item
  
  Structures can only be assembled in areas where energy density is high
  enough to support such assembly.
  
\item
  
  There are rules defining the space of possible structures available
  for assembly by our enerstatic loop

  \begin{enumerate}
  \def\labelenumii{\alph{enumii}.}
  \item
    
    A structure is minimally composed of the following elements

    \begin{enumerate}
    \def\labelenumiii{\roman{enumiii}.}
    \item
      
      Cost of assembling structure
      
    \item
      
      Cost of dissembling structure
      
    \item
      
      Position in space
      
    \item
      
      Causal Power of the structure
      
    \item
      
      Cost of the causal power

      \begin{enumerate}
      \def\labelenumiv{\arabic{enumiv}.}
      \item
        
        Causal power always has a cost and ``eats up'' energy based on
        the particular effect that the power has at every time step. Not
        meeting this energy requirement would result in the
        non-existence of the structure.
        
      \end{enumerate}
    \end{enumerate}
  \item
    
    Causal power of a structure consists of the following properties:

    \begin{enumerate}
    \def\labelenumiii{\roman{enumiii}.}
    \item
      
      Sensed Input Properties, this is the set of properties that the
      structure is ``aware'' of.
      
    \item
      
      Affected Output Properties, this is the set of properties that the
      structure exerts its causal power on. i.e.. properties that are
      modified by the structure.
      
    \item
      
      The effective radius, this is the distance at which properties can
      be sensed, or affected. This may be different for different
      properties.
      
    \end{enumerate}
  \end{enumerate}
\item
  
  When the system is unable to perform actions that allow for an
  appropriate flow of energy, the system ``dies'' i.e. falls into
  equilibrium.
  
\end{enumerate}

To get a better idea of how to meet all these guidelines, we will
quickly fashion together a minimal enerstatic network consisting of
minimal enerstatic loops. To begin with, we acknowledge that an
enerstatic loop is a control system, which formally consists of four
components: (1) A ``control center'', which defines the ideal ``set
point'', or state, of the system (2) A ``physiological variable'', or a
``sense'' of the variable to be regulated(3) ``Sensors'', which are used
to gain information about the environment (4) ``Effectors'', which are
mechanisms used to modify the environment. This takes us well on our way
to meeting all our previous guidelines. The following is working
javascript code for the most minimal enerstatic loop, where we can see
all of these features reflected:

{\centering
  \includegraphics[height=5in]{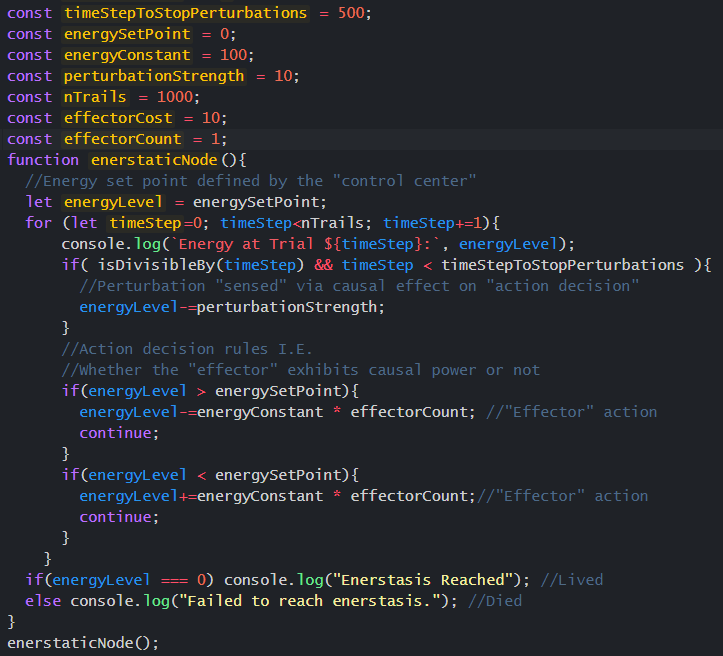}
\par}

In our case, the ``Set point'', or ``Control Center'' is 0 and
represents energy, in other words, the energy level that the enerstatic
loop wants to maintain is 0. Therefore, this value is both the setpoint
and the physiological variable to be regulated. Our minimal enerstatic
loop only has one sensor, and that is a sense of its physiological
variable, its energy. It is implicitly a sense in that when the energy
level changes, this has an effect on whether the ``effectors'' it has
exhibit their causal power or not. The loop has a ``sense'' of its
environment through the effect that the environment has on its
physiological environment, and reacts as to remove the effects of this
energetic perturbation.

{
  \centering
  \includegraphics[width=3.73737in,height=3.57208in]{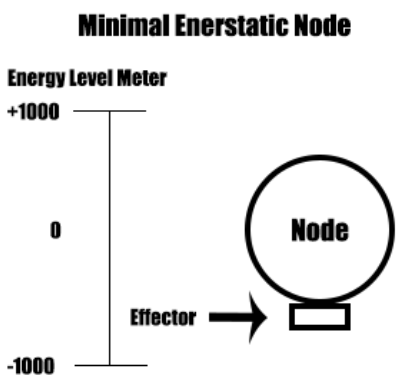}
  \par
}

Our minimal enerstatic loop is lacking quite a few things to make it
interesting/extensible, specifically:

\begin{enumerate}
\def\labelenumi{\arabic{enumi}.}
\item
  
  There is only one loop in the environment
  
\item
  
  Its structure (i.e. its single effector) is fixed in that it cannot be
  added/removed.
  
\item
  
  The single effector doesn't have an assembly/disassembly and causal
  power cost
  
\item
  
  Death of the loop isn't explicitly defined
  
\item
  
  Cannot take active actions, and doesn't have any sort of learning
  capacity
  
\item
  
  The flow of energy into the system doesn't depend on the actions of
  that system
  
\item
  
  There no loss of energy as the loop operates
  
\end{enumerate}

{\centering
  \includegraphics[width=5.25521in,height=3.03185in]{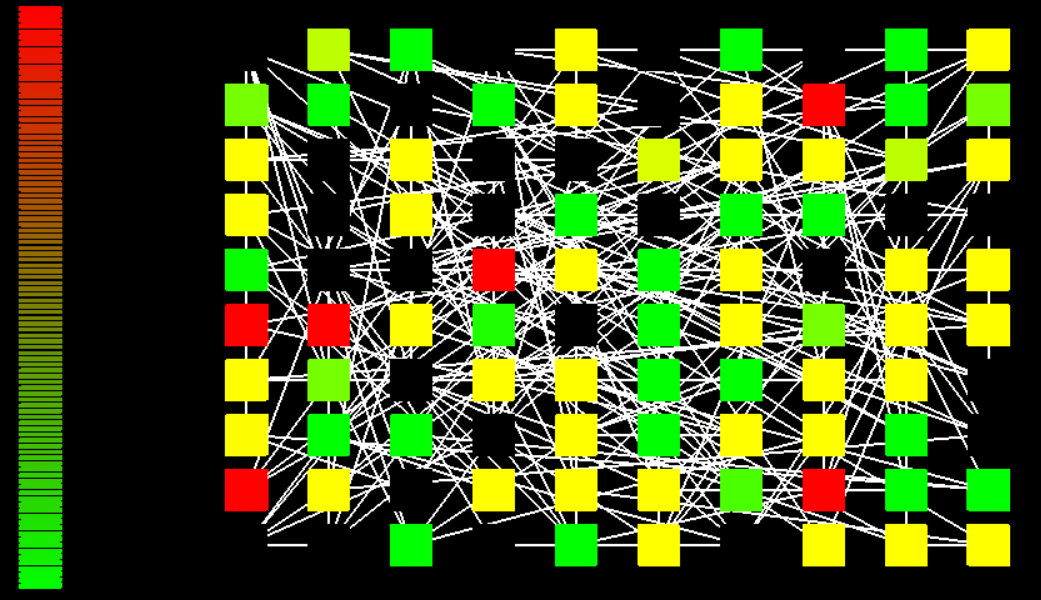}
\par}

\begin{itemize}
\item
  
  Example of a Minimal Enerstatic Network (MEN): Loop health is
  represented according to the color of the loop. Towards red means that
  a loop is close to its fatal energy threshold, whether that's too
  much, or too little energy. The opacity/transparency of each loop
  moves towards clear when the loop is towards its negative energy
  threshold. Dead loops end up invisible, but their associated energy
  channels do not. Each loop has 3 instances of a structure called an
  ``Energy Channel'', whose causal powers shifts energy between origin
  and target loops based on the origin\textquotesingle s energy level.
  Thus, each loop is randomly connected to 3 other loops. The network
  can then be perturbed in various ways to see whether or not it is able
  to adapt.
  
\end{itemize}

Addressing the specific code implementation of all these details is
outside the scope of this paper, but javascript code implementing these
aspects (used to generate the previous image), is available here:
\href{https://github.com/Tyfoods/minimal-enerstatic-network}{\uline{https://github.com/Tyfoods/minimal-enerstatic-network}}

At this point, there are a couple guidelines that may need a bit more
clarity with respect to their motivation. The first is that idea that an
enerstatic loop has a set of structures which it can assemble, and
disassemble and the second is the nature of the structures themselves.
The organizational nature of an enerstatic loop is meant to capture the
idea from assembly theory that information about how to construct
particular structures is embodied by what we would call living things.
This information embodiment is what allows assembly theory to
characterize life as that which produces a high copy of structures above
a particular complexity threshold, and furthermore to place different
living things on different points on the ``life continuum''. Structures
are an abstraction of enerstatic loops that allow one to specify the
function of such loops without referring to the various control system
details that we have previously discussed. We are allowed to do this
because these structures represent enerstatic loops that are kept
consistently at their homeostatic set point and thus exert the same
causal power consistently. We take the assumption that the enerstatic
loop that these structures are a part of considers structures outside of
homeostasis to be ``defunct'' and thus disassembles them. What will be
discussed next is what a learning algorithm could look like in an
enerstatic network.

\hypertarget{learning-in-enerstatic-networks}{%
\subsection{Learning in Enerstatic
Networks}\label{learning-in-enerstatic-networks}}

In our minimal enerstatic network, each loop was extremely sensitive to
any perturbations in its energy level. In other words, any fluctuation
in internal energy resulted in ``action'' being exercised by the loop.
In our case, this action was in the form of causal power exhibited by
our single effector. There is another sort of action that can take
place, and this action is with respect to the reconfiguration of the
enerstatic loop. If we allow our enerstatic loop to reconfigure itself,
and give it time to test the consequences of its new architecture, then
we can employ some sort of learning algorithm.

A great way to give enerstatic loops time to test the consequences of
their actions is through implementing two ``windows'', or ranges, inside
of its allowed energy levels. These two windows are the ``stasis
Window'', and the ``action Window''. The stasis window is a set of
numbers around the energy set point of the enerstatic window in which
the loop's architecture goes unchanged. The size of the window defines
how much time, or tolerance, a loop has before it takes the action of
reconfiguring its architecture. The action window is a set of numbers
around the positive and negative bounds of the stasis window which
indicate that the loop should attempt to reconfigure its architecture.

{

  \centering
  \includegraphics[width=4.21354in,height=5.1303in]{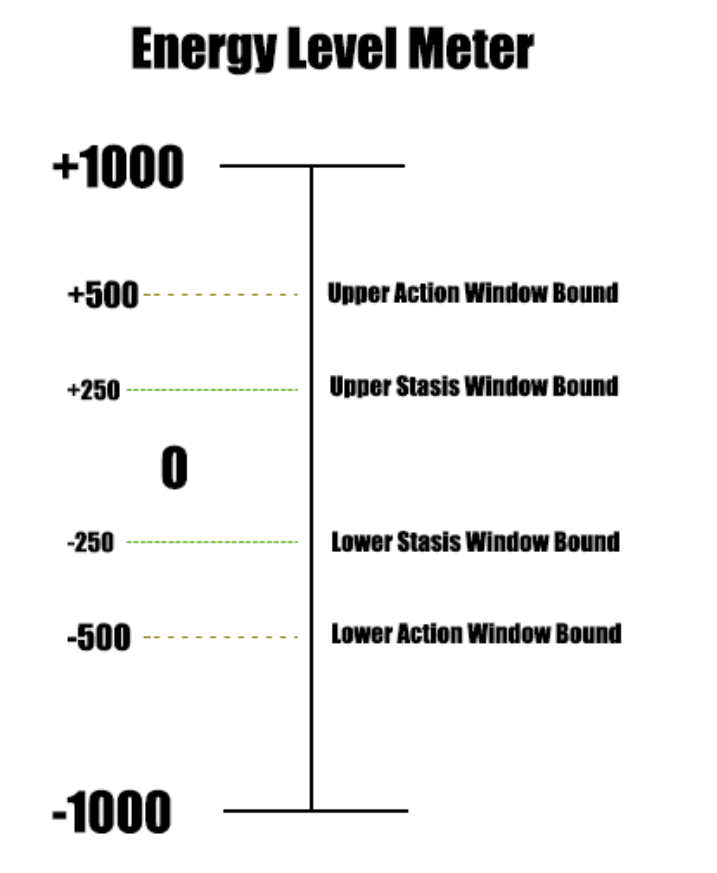}
  \par
}

\begin{itemize}
\item
  
  Enerstatic window parsed into two separate windows: The stasis window
  in which no action is taken, thus the loop is considered stable, and
  the action window, in which the loop is considered unstable, and must
  take action to return to a stable state.
  
\end{itemize}

Now that we know when we will reconfigure the enerstatic loop, we are
now tasked with defining under what particular modification/action will
be made, how many of those modifications will be made and under what
circumstances. A simple approach is to define a probability associated
with each action, and to allow for only one action to be made per time
step. In other words, when the loop is pushed into the action window,
each possible action has a probability associated with it, which
determines whether or not that action is the action selected on that
time step. However, it is not necessarily true that an action MUST be
taken on any given time step if a loop is in its action window. This can
be left up to the programmer and isn't important for our conceptual
discussion here.

The question the programmer must ask next is: ``When and how should I
update my beliefs, or probabilities?'' In this example, we answer both
of these questions by introducing yet another window, the ``Change
Action Probability'' (CAP) window. Collectively, these windows are known
as enerstatic windows. The CAP window allows for probabilities to be
changed in a sort of last ditch effort to attain stasis. Since the goal
is to change action probabilities in such a way that result in a return
to stasis, it would be reasonable to come up with a complicated
algorithm that attempts to predict what actions should be modified,
however, inspired by plasticity mechanisms in neurons, we can opt for
simple associative hebbian-like learning. During the action window, the
loop can integrate the information, or series of steps that led to the
loop being pushed into the CAP window. In particular, the loop can be
made to remember the actions that were taken as well as what the energy
level was when each action was taken. Instead of looking any deeper at
the details of this data, our simple associative hebbian-like learning
algorithm strengthens, or weakens all actions taken in the action window
based on whether those actions led to the stasis window, or the CAP
window. In this way, probabilistic sequences of actions that helped are
strengthened, while probabilistic sequences of actions that hurt, are
weakened. Furthermore, the particular energy level at which the action
was taken could be recorded to allow for an even more detailed belief
updating.

While exploring the space of possible enerstatic networks is an
interesting, and potentially valuable, task in its own right; Modeling
physical phenomena is a powerful way to both gain insight into existing
SOULS as well as build new even more heavily nature-inspired SOULS. As
an example, we extend the concept of an enerstatic network to model the
function of neurons.

\hypertarget{modeling-neurons-enerstatic-neural-networks}{%
\subsection{Modeling Neurons: Enerstatic Neural
Networks}\label{modeling-neurons-enerstatic-neural-networks}}

Before conceptually detailing how one might go about modeling neurons
with enerstatic networks, it is important to understand the motivation
behind the approach in the context of biology because it may apply
generally to all cells. The goal of modeling cells as enerstatic
networks is to investigate if such an idea can serve as a guiding
principle to understanding the behavior of cellular collectives. In the
particular case of neurons, understanding brains. Neurons were the main
inspiration in the development of the enerstatic network due to the
observation that neurons are, for the most part, lifelong cells that
collectively take up 20\% of the body\textquotesingle s energy budget
\textbf{\cite{68, 69}}. The question that this provoked was: ``Given that we see
neurons that exist over the lifetime of an organism, what would that
neuron's behavior have to be like in order for it to do so?'' This paper
suggests that the answer is ``behavior which regulates the neuron's
energy level within a certain homeostatic window''.

Indeed a neuron with low enough levels of ATP will not even be able to
carry out apoptosis and instead will undergo necrosis \textbf{\cite{56}}. On the
contrary, high levels of ATP in a neuron can facilitate cytotoxicity and
ultimately cause cell death \textbf{\cite{57}}, or cause disruptions in protein
synthesis, which can also lead to cell death \textbf{\cite{58}}. This demonstrates
the importance of ATP as the ``physiological variable'' that the
``control system'' that is the neuron, regulates.

If the primary objective of a neuron is to control its energy level,
then one could say that ``information processing'' in a neuron is
``built on top'' of the mechanisms that a neuron uses to regulate its
energy level. In this case, it would be the case that trying to
understand the way neurons process information would be frustrated by
the fact that neurons are energy management systems first, and
information processors second. There is indeed evidence demonstrating an
informative relationship between energy efficiency and function
\textbf{\cite{54, 55}}. More generally, if all cells that exist must meet
this existential imperative, then it could often make sense to model and
understand them with enerstatic networks. In this perspective, all cells
don't know what ``computations'' they're involved in, or what their role
is in the development and operation of the organism they are embedded
in. Cells only try to exist for as long as possible given their
environment. The purpose of the structures that cells assemble and
maintain is primarily to allow those cells to maintain their energy
within a homeostatic window. This frustration is indeed one and the same
as that which is caused by the ``Behavioral illusion'' discussed in
``IOS Illusions: Motivating Modeling With Enerstatic Loops''. We now
turn to an example of using enerstatic networks to model neurons.

Since computational experiments with enerstatic neural networks are
currently in progress, and the purpose of this section is only to
inspire ways of modeling biological SOULS with enerstatic networks, only
the general conceptual details of the project will be discussed.

Enerstatic neural networks are just minimal enerstatic networks except
for the fact that the structures that loops in an enerstatic neural
network assemble are analogous to those in neurons to the programmers
desired level of complexity. Using the requirements for an enerstatic
network, we can specify our conceptual model as follows:

\begin{enumerate}
  \def\labelenumi{\arabic{enumi}.}

  \item{There is a flow of energy into the system that depends on the action of that system
    \begin{itemize}
      \item{Neurons receive blood flow, or energy, only when they are stimulated. This is inspired by neurovascular coupling (NVC).}
    \end{itemize}
  }

  \item{This energy flow results in the establishment of an energy distribution, where energy is higher and lower in different parts of the system in the form of other enerstatic loops, or ``structures''. The establishment of this energy distribution represents the system starting out ``alive'' I.E. in a non-equilibrium steady state.
    \begin{itemize}
      \item{Each neuron starts out with a suite of structures such as ion channels, receptors, axons, dendrites, genes etc.... Each structure occupies a different point in space and costs a variable amount of energy to create.}
    \end{itemize}
  }

  \item{This energy flow results in the aforementioned internal energy distribution being modified in some way
    \begin{itemize}
      \item{Energy is captured and released by mitochondria which despite their motility, are unequally distributed about the neuron. This results in local pockets of high and low energy.}
    \end{itemize}
  }

  \item{There is an inevitable loss of energy as the system generates entropy
    \begin{itemize}
      \item{Neurons have a “causal power” cost that is defined as a function of the structures they’ve assembled.}
    \end{itemize}
  }

  \item{Structures can only be assembled at these energetic points
    \begin{itemize}
      \item{Only local areas with high enough energy are capable of assembling structures.}
    \end{itemize}
  }

  \item{There are rules defining the space of possible structures.
    \begin{itemize}
      \item{As mentioned during our minimal enerstatic loop exercise, the cost of structure assembly, disassembly, and causal power are up to the programmer, but can be determined through empirical measurements. Similarly, conditions under which a structure exerts its causal power can be determined either way. Thus, for brevity, and because it is besides the point for this example, specific values/conditions have been left out.}

      \item{\textbf{Structures}
        
        \begin{itemize}
    
          \item{\textbf{Macrostructures}
            \begin{enumerate}
              \item[1.]{Genes - It is important to note that a genome doesn’t necessarily specify all possible structures in the space of our program.
                \begin{itemize}
                   \item[\textbullet]{\textbf{Structural Genes} - Produces a particular “physical” structure which modifies the loop's architecture.}
                   \item[\textbullet]{\textbf{Regulatory Genes} - Specifies how gene expression (causal power exertion) in one gene affects the expression of other genes.}
                \end{itemize}
              }
  
              \item[2.]{Soma
                \begin{itemize}
                   \item[\textbullet]{Geometrically represented as a sphere with some diameter specified by the genes}
                   \item[\textbullet]{Membrane potential specified by genes}
                \end{itemize}
              }
  
              \item[3.]{Axons
                \begin{itemize}
                   \item[\textbullet]{Geometrically represented as a 2 or 3D lines}
                   \item[\textbullet]{Membrane potential specified by genes}
                \end{itemize}
              }
  
              \item[4.]{Dendrites
                \begin{itemize}
                   \item[\textbullet]{Geometrically represented as 2 or 3D lines}
                   \item[\textbullet]{Membrane potential specified by genes}
                \end{itemize}
              }
            \end{enumerate}
          }

          \item{\textbf{Microstructures}
            \begin{enumerate}
              \item[1.]{Voltage Gated Ion Channels - Voltage sensitivity specified by genes. Activates when the membrane potential of the structure it is embedded in reaches a certain threshold. When activated, either depolarizes (raises), or hyperpolarizes (lowers) the membrane potential of the cell; the degree to which this lowering happens is specified by the genes.}
  
              \item[2.]{Ligand Gated Ion Channels - Activated when in contact with some particular neurotransmitter specified by the genes. When activated, either depolarizes (raises), or hyperpolarizes (lowers) the membrane potential of the cell.}

              \item[3.]{ATPase pumps - Consume a certain amount of energy per time step thus causing local energy demand.}

              \item[4.]{Neurotransmitters - Upon contact, activates particular ligand gated ion channels as specified by the genes.}

              \item[5.]{Mitochondria
                \begin{itemize}
                   \item[\textbullet]{Moves around according to energy gradients, preferentially going where energy is low.}
                   \item[\textbullet]{Receives energy upon receiving blood flow and outputs energy to the local area.}
                \end{itemize}
              }

              \item[6.]{mRNA - Instructions sent out by the nucleus to ribosomes which detail which structure to assemble.}

              \item[7.]{Ribosomes
                \begin{itemize}
                   \item[\textbullet]{Moves according to energy gradients, preferentially going where energy is high.}
                   \item[\textbullet]{Upon contact, and given sufficient local energy, translates mRNA into a structure.}
                \end{itemize}
              }

            \end{enumerate}
          }

        \end{itemize}
      }
    
    \end{itemize}
  }

  \item{When the system is unable to perform actions that allow for an appropriate flow of energy, the system dies I.E. falls into equilibrium.
    \begin{itemize}
      \item{When neurons hit their fatal energy threshold, they undergo apoptosis and shut down shop. In particular, all their associated structures are removed from the program.}
    \end{itemize}
  }
  
\end{enumerate}

{\centering
  \includegraphics[width=2.55952in,height=3.35938in]{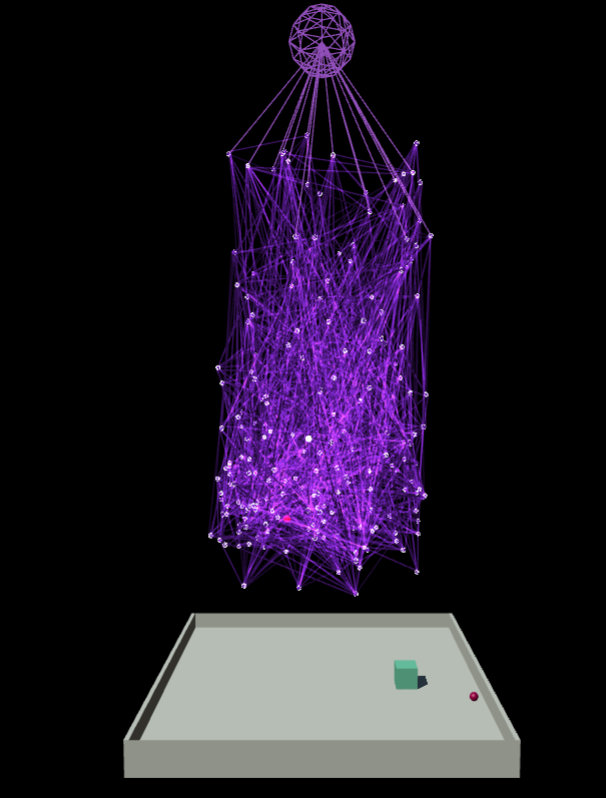}
\par}

\begin{itemize}
\item
  
  Visualization of an ongoing experiment featuring an enerstatic neural
  network receiving visual inputs in the form of a ``Sparse Distributed
  Representation (SDR)'' \textbf{\cite{50}}. The enerstatic neural
  network must control the ball and the camera embedded in the ball to
  accomplish two things: (1) Recognize food (2) Move towards food --
  Otherwise, it dies. This makes enerstatic networks a great candidate
  for open-ended evolution.
  
\end{itemize}

As mentioned, one can be as biophysically detailed with model parameters
as they'd like. This may result in important insights as to the inner
workings of their actual biological counterparts. However, when
discussing the creation of technology, it has been evidenced by advances
in artificial intelligence, that one needn\textquotesingle t bother with
setting all parameters, and structural details, as accurately as
possible to yield something that is both technically useful, and worthy
of scientific inquiry. Indeed, despite being dramatically simpler than
real biological neural networks, Artificial Neural Networks have been
used to gain insights into their behavior. Modeling physical phenomena
with enerstatic networks thus potentially provides a way to both
understand those phenomena and use the resulting technology to do
something useful.

An interesting way to use enerstatic neural network models to generate
more biologically relevant results, might be to map abstraction layers
onto each other. For example, a Hodgkin Huxley Neuron (HHN) is less
complicated than a biological neuron (BN), but empirical data can be
used to fit a HHN to a BN. Similarly, one could then use empirical data
generated by the HHN neuron to map onto yet an even simpler neuron
model. This may allow one\textquotesingle s model to be much more
relevant to experimental inquiry. Such abstraction layer mapping is part
of the ongoing computational experiments involving enerstatic neural
networks.

Further ongoing experiments are in the domain of applying open-ended
evolutionary techniques to enerstatic networks in general. The advantage
of using enerstatic networks as a tool for building technologies is its
natural symbiosis with such techniques. As previously mentioned,
open-ended evolution requires building blocks with a particular kind of
dynamics which allow for an increase in complexity through the
construction of ever larger building blocks as selected for by an
intelligence which is itself constructed and amplified from within the
system. This selection mechanism is built into enerstatic loops by
default making them great candidates for open-ended evolution.

\hypertarget{evolving-enerstatic-networks}{%
\section{Evolving Enerstatic
Networks}\label{evolving-enerstatic-networks}}

The key properties that make enerstatic loops, and therefore enerstatic
networks, symbiotic with open ended evolutionary techniques can be
summed up in the following way: Enerstatic loops can only persist their
existence if they are able to modify their architecture in real-time in
such a way that they take in more energy than they lose.

If the only way that enerstatic loops can receive energy is by
successfully organizing themselves in particular ways based on the
environment that they are embedded in, then there is no need to specify
particular rewards for particular tasks because such rewards are
intrinsic to their nature. With currently available tools, this makes it
difficult to engineer an enerstatic network to do exactly what you'd
like it to do, but it does open the doors of opportunity to evolving
networks that both achieve your desired behavior and exhibit behaviors
you may have never come up with yourself.

The goal of open-ended evolution is to generate seemingly never-ending
complexity as we observe in our universe. The challenge of developing a
system with such behavior is centered around the environment-agent
dilemma. The dilemma is that, as an agent embedded inside some
environment becomes more complex, or intelligent. It essentially masters
the environment thereby maxing out its own complexity and intelligence.
Thus, arguments have been made for creating environments that complexify
along with the agent that is embedded inside of it \textbf{\cite{51, 52}}.
While seemingly straightforward, the problem of how to continuously
complexify the environment along with the agent, without periodic
injection of some outside intelligence (I.E. programmer intervention),
has been elusive.

A large part of the problem is how the encoding of the environment
operates. Although, there exist universal encodings for certain domains
like mazes, or 2D landscapes. The problem with expanding beyond
particular environmental domains seems to persist. In their paper, the
authors of Evolution through Large Models (ELM) propose that: ``computer
programs are a general and powerful encoding for continually expanding
the richness of an existing environment.'' and takes a critical step
towards demonstrating how use of raw computer programs could be used to
ensure open-endedness indefinitely \textbf{\cite{9}}. In their paper,
they develop a particular methodology that results in a program that is
able to conditionally output computer programs which they call an
``invention pipeline''. In short, the methodology consists of three
steps:

\begin{enumerate}
\def\labelenumi{\arabic{enumi}.}
\item
  
  Use any arbitrary large language model (LLM) to provide intelligent
  mutations to an evolutionary algorithm (In this case, the quality
  diversity (QD) algorithm, MAP-Elites) to generate training data in a
  domain where little, or none, exists.
  
\item
  
  Pre-train the LLM with aforementioned training data, such that it can
  now output similar data
  
\item
  
  Use reinforcement learning, and some additional tweaking, to fine tune
  the LLM such that it is able to produce data based on the condition
  that it is exposed to.
  
\end{enumerate}

This methodology captures a few key elements that we believe are key to
indefinite open-endedness, which are as follows:

\begin{enumerate}
\def\labelenumi{\arabic{enumi}.}
\item
  
  Increasingly intelligent (correlated) mutations over indirect
  encodings, which are nonetheless, just code.
  
\item
  
  Generation of a diverse amount of novel structures without the use of
  a trained neural network.
  
\item
  
  The creation of an agent that is able to create structures
  conditionally based on its ``needs''.
  
\end{enumerate}

Although different in form, It is these elements that evolutionary
enerstatic networks aim to achieve.

\hypertarget{energetically-closed-systems}{%
\subsection{Energetically closed
systems}\label{energetically-closed-systems}}

Fundamentally, enerstatic networks are meant to capture the idea that
any ``thing'' that exists is an ``energetically open system''. An
energetically open system is a system that takes as input some finite
amount of energy, and outputs only a fraction of that amount of energy.
Specifically our energetic open systems consisted of an extension of
these three rules:

\begin{enumerate}
\def\labelenumi{\arabic{enumi}.}
\item
  
  There must be a source of energy flowing into the system
  
\item
  
  There must be mechanisms that essentially trap energy inside the
  system.
  
\item
  
  There must be mechanisms responsible for energy flowing out of the
  system
  
\end{enumerate}

In this setting, we had a rather static environment that was not capable
of becoming much more complex in response to the complexity of our
enerstatic network. However, we have mentioned that as a consequence of
the information symmetry of the free energy principle, the environment
must also be treated as an agent capable of, in a sense, increasing its
complexity \textbf{\cite{18}}. One way to realize this information
symmetry is to treat the environment itself as an enerstatic loop. In
this case, we now have an ``energetically closed system''. This paper
suggests that using such an approach, along with methods like ELM, is
critical to the development of truly open-ended processes. Our
energetically closed system follows these rules:

\begin{itemize}
\item
  
  Some finite amount of energy exists inside the system
  
\item
  
  Energy cannot be created nor destroyed
  
\item
  
  Energetically open system rules previously specified apply to all
  structures within the system* (We'll later see a small exception, but
  it's mostly correct).
  
\item
  
  Environment is modeled as an enerstatic loop that minimizes free
  energy (I.E. the energy not currently trapped within a structure)
  
\item
  
  Environmental enerstatic loop (EEL) is responsible for building
  structures within it
  
\end{itemize}

In order to get a better grip on the approach being developed here, it
is best to run a sort of simulation, or thought experiment, over what we
could feasibly do within the bounds of our required specifications
interspersed with motivations for why we might choose to do things in
that particular way. Of course, there are many particular ways, much of
which can be co-opted for application to energetically open systems.
Following the theme of this paper: The following is a thought experiment
whose iterations are currently in development -- Consider this to be
``food for thought'' rather than a definitive ``how to'' guide.

\hypertarget{causal-niches}{%
\subsubsection{Causal Niches}\label{causal-niches}}

At first, we only have an environmental enerstatic loop (EEL), whose
"goal" is to minimize free energy. Since all energy is free at first,
the first thing the loop "wants" to do is trap energy within structures.
The first structure that it generates, S\textsubscript{0}, must
necessarily have causal power such that the environment recreates it. To
actually implement this, we will require that our EEL has a property
which details how much energy each structure inside it should be
receiving each timestep. The default amount of energy is 0. Thus, in
order to exist S\textsubscript{0} must sense this particular property
and set it equal to its requisite free energy demand (FED), or higher.
There are three scenarios here for any given structure:

\begin{enumerate}
\def\labelenumi{\arabic{enumi}.}
\item
  
  If the structure receives less than its FED, it ceases to exist.
  
\item
  
  If the structure receives exactly its FED, it persists its existence.
  
\item
  
  If the structure receives more than its FED, and this energy is not
  explicitly used, then this energy remains trapped unless explicitly
  moved by some other ``thing''.

  \begin{enumerate}
  \def\labelenumii{\alph{enumii}.}
  \item
    
    Importantly, the structure must receive less than its free energy
    limit (FEL), which places a bound on the amount of energy a
    structure can contain.
    
  \end{enumerate}
\end{enumerate}

Recall that this dynamic is in alignment with our aforementioned
reasoning with respect to structures inside of an enerstatic loop.
Structures inside a loop are abstracted enerstatic loops that are kept
consistently at their homeostatic set point and thus exert the same
causal power consistently. We take the assumption that the enerstatic
loop that these structures are a part of consider structures outside of
homeostasis to be ``defunkt'' and thus disassemble them.

Importantly, S\textsubscript{0} is \textbf{only} capable of interacting
with EEL and no other structure (S\textsubscript{n}). This is due to the
fact that it is not possible to explicitly specify an interaction of
S\textsubscript{0} with the unknown properties of a yet non-existent
structure. However, S\textsubscript{0} can interact with future
structures indirectly through its effect on the environment. This
restriction of only being able to interact with structures that have
been previously generated remains true for every subsequent structure,
which gives us a graph illustrating the causal niche of each generated
structure. A causal niche is defined as the set of properties that a
structure can sense and affect.

{
  \centering
  \includegraphics[width=4.32813in,height=4.1894in]{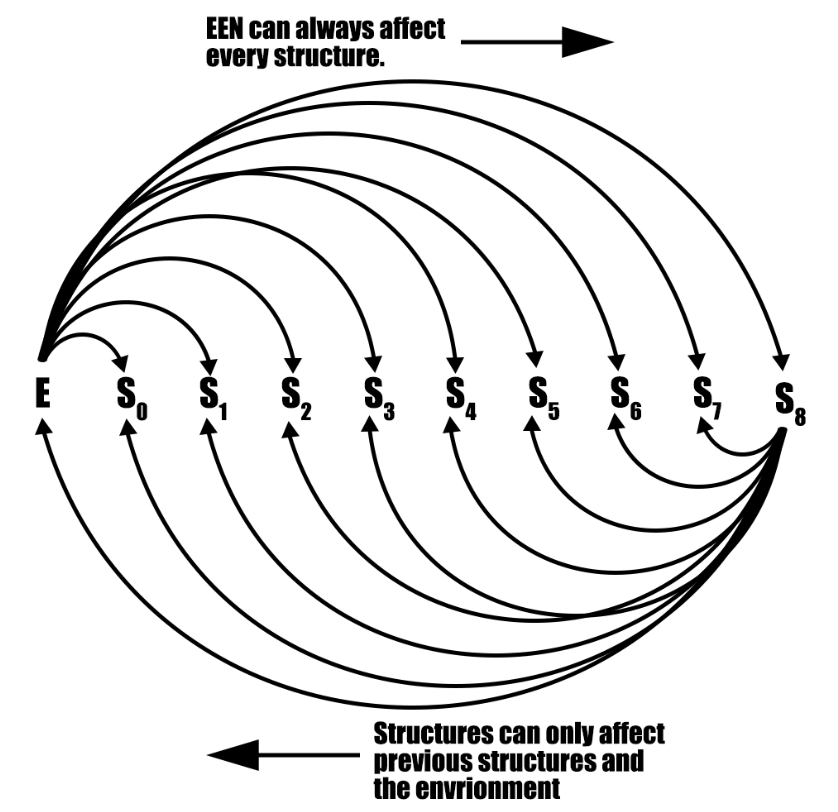}
\par}

\begin{itemize}
\item
  
  Causal Niche Graph demonstrating that structures can only affect the
  environment and previous structures that have been created.
  Illustrated here are the arrows for structure 8 and the environment,
  which can always affect every structure.
  
\end{itemize}

The reason that structures are defined like this is because the
alternative is intractable, or at minimum, much closer to the practice
of inducting the laws of physics. The reasoning is as follows: In order
to generate a structure that is capable of explicitly affecting a future
structure, one needs to have a medium (i.e. universal method of
communication) that is able to underwrite those particular
specifications. This medium is what every structure ``lives in'' and
operates through. A good analogy for better understanding is with the
case of a physics engine. In a physics engine, we are typically only
concerned with Newtonian physics, in other words, with the question of
``What happens when object A bumps into object B?''. Since the
underlying physics are known, one doesn't need to worry about which
object was created first in order to have the objects interact properly.
The laws of physics that detail how forces act on objects in order to
move them, are known. In our case, the laws of physics are not known,
and thus the interaction between two ``objects'' need to be explicitly
defined via code which defines the object\textquotesingle s causal
power.

The requirement of explicitly specified interactions is important for
explainability. There have indeed been efforts for exploring possible
open-ended evolutionary systems using basic universal rules which
constitute the system's ``laws of physics''. Three notable systems are
n-dimensional discrete cellular automata, continuous cellular automata,
and hypergraphs. The issue here is that, not only can we not guarantee
open-ended evolution, but that even if we managed to achieve it, it
would be difficult to understand what is happening in a way that is
human-interpretable. Therefore, using explicitly specified interactions
via code, despite such a restriction, will be important for
understanding both what is going on, and potentially, for how to inch
closer to open-ended evolution in a principled manner.

{
\centering
  \includegraphics[width=2.99479in,height=2.55208in]{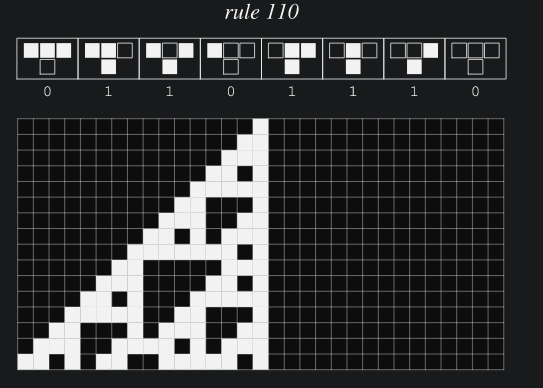}
\includegraphics[width=2.96875in,height=3in]{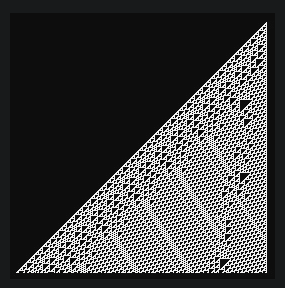}
\par
}

\begin{itemize}
\item
  1D cellular automata called ``Rule 110'' demonstrating the amount of
  complexity one can achieve from running a simple rule over a simple
  initial condition. Despite starting with a single black cell and a
  simple rule for determining how each cell changes color (It does so
  only based on its immediate neighbors), one ends up with
  unpredictable, and potentially never-ending, complexity. In other
  words, simple rules can give rise to arbitrarily complex behavior.
  Indeed, rule 110 was found to be capable of ``universal computation'',
  in other words, like a typical computer, you can use rule 110 to run
  any possible computer program. The problem here is the difficulty in
  encoding and interpreting the program. Interestingly, simply exploring
  the space of possible simple programs has revealed patterns that are
  similar, at a macro-level, to the kinds that have been found in
  nature. As a related note, Stephen Wolfram heads a research program
  focused on discovering a fundamental theory of physics by exploring
  the space of possible simple hypergraph programs, which are
  essentially even simpler versions of cellular automata. Through this
  exploration, Wolfram Research has been able to mathematically derive
  essential features of special relativity from hypergraphs, such as
  Lorentz transformations and the speed of light as a universal speed
  limit \textbf{\cite{70}}
  
\item
  On the left, the rules of the program and the first 16 steps, on the
  right, the first 250 steps. We start from a single black cell on the
  first row and apply the rule to each cell in the row to generate the
  next row.
  
\item
  Source: Weisstein, Eric W. "Rule 110." From MathWorld-\/-A Wolfram Web
  Resource.
  \href{https://mathworld.wolfram.com/Rule110.html}{\nolinkurl{https://mathworld.wolfram.com/Rule110.html}}
  
\end{itemize}

\hypertarget{structure-specifications}{%
\subsubsection{Structure
specifications}\label{structure-specifications}}

As illustrated in our causal niche graph, we have a total of 9
structures generated, but they're completely abstract. All we know is
that the minimum requirement for each structure is that there must be a
cost of assembling/disassembling, causal power (and its associated
cost), effective radius and a position that they occupy in space. There
are other properties that one can apply to these structures, which could
either be ubiquitous to all structures, or defined uniquely only to some
structures.

In the case of ubiquitous properties, the thought is that these should
be directly computable in some way from the aforementioned properties so
that they remain related to the specifications of that structure. For
example, the mass of a structure could be calculated based on the
energetic cost of the structure using E = mc\textsuperscript{2} →
E/c\textsuperscript{2} = m, where E = energy, m = mass, and c = constant
speed of light, or cosmic speed limit. Furthermore, position could then
be modified as a function of mass, where the position of an object
requires a certain amount of energy to be perturbed. This could be
calculated using: W = (1/2) * m * v\^{}2 where W is the work done on the
object, m is the mass of the object, and v is its final velocity. The
amount of energy required to perturb a property's value by ``1'' unit is
here called the ``perturbation energy'' and is important for
energetically grounding any property whether it is unique, or
ubiquitous.

An example of a unique property might be ``membrane voltage''. The
property should minimally have a default value, and a perturbation
energy, however a third likely important characteristic is a
``location'' within the causal power of the structure. This will allow
the causal power of a structure to be modified through perturbations of
structure properties. One could imagine utilizing current membrane
voltage to decide whether to send a message, or not. For this reason,
we'll refer to such properties as ``causal properties''. Similar to how
equations that have been empirically discovered were used to decide how
to modify the position of any structure, equations which describe how
difficult it is to modify membrane voltage could be implemented. Thus,
such properties can assist in generating an energetically closed system
that is not just interpretable, but also bears semblance to our physical
reality thus potentially serving as a powerful modeling tool. However,
it must be noted that this is not without the cost of dramatically
increasing the complexity of implementation. As for actually defining
these properties, this is something we'd like to leave to an automated
task.

\hypertarget{automatic-structure-generation}{%
\subsubsection{Automatic structure
generation}\label{automatic-structure-generation}}

The causal niche graph allows us to generate structures continuously in
a principled way. We can fill niches with various kinds of structures in
each causal niche. Since each causal niche restricts only the sensed
input properties and affected output properties, the actual causal
dynamic that implements the sensed information and affects the output
properties is up to us. To offload this work, we take a hint from ELMs
usage of Large Language Models (LLMs). In effect, we've created a
function that has certain inputs (sensed set of properties), potential
internal variables (causal properties), and outputs (affected set of
properties), and we'd like to use a Large Language Model to fill in the
details about how inputs and internal variables produce on output.
Utilizing LLMs, the environment can create a set of structures with
various causal dynamics in every niche -- The limit of structures within
a niche is up to the programmer. As for the effective radius, the LLM
can be used, or the programmer can opt to assign effective radii based
on some particular algorithm or criteria.

In addition to arguing that the goal of the environment is to minimize
its free energy, this paper also suggests that one good way to do this
is for the environment to actually test different sets of structures
concurrently to see which sets of structures combine to make complex
composite structures capable of minimizing VFE. Those that fail to
minimize VFE well enough get outcompeted and thus end up with a lower
likelihood of being created in the future.

Before discussing how structures can mix and match in this way, it is
important to discuss a mechanism for automatically assigning energy
costs to such structures.

\hypertarget{automatic-energy-cost-assignment}{%
\subsubsection{Automatic energy cost
assignment}\label{automatic-energy-cost-assignment}}

Previously, we assigned energy costs to structures either using
intuition or empirical data from experimentation in the service of
modeling. The caveat is that one needs to be careful not to violate the
constraint that energy cannot be created or destroyed. With this in
mind, it is possible to do something similar to what we did before, but
we can also opt for a different method that allows for automation. The
benefit of assigning energy costs automatically is not just to avoid the
work of doing so, but also to allow for energetic consistency throughout
various created structures. In other words, similar causal powers have
similar energetic costs. This provides us with a sort of ``underlying
physics'' that our enerstatic program must obey. It should be noted that
this same technique can be applied to energetically open systems of
enerstatic networks as well, but may be less straightforward to apply in
the context of modeling.

In order to automatically assign energy costs to structures, we first
make the claim that the cost of structure assembly and disassembly are
equal and opposite. Furthermore, the cost of structure assembly is equal
to the cost of causal power. Since causal power is just a function whose
body is code that has been generated by an LLM, we can take that code
and convert it into an ``Abstract Syntax Tree'' (AST). An AST is a
simplified representation of a piece of code that is typically
constructed by a compiler, or interpreter as a way of efficiently
analyzing and processing source code. Technical details aside, the AST
allows us to calculate the number of fundamental operations, or
primitives, that are inside of the code defining our causal dynamic.
With knowledge of all the possible operations in our programming
language of choice, we can assign energy costs to each of these
operations and then calculate the total amount of energy that our causal
dynamic costs. Toggling these costs among inside of, and between,
different programming languages opens up an interesting domain of study.
As for causal properties, the cost of these must be calculated using
their perturbation energy function. Such a function receives the current
value of the property and returns the amount of energy required to
modify that value. One can then calculate the cost of the property by
starting at 0 and iteratively summing the costs as it is raised to the
default value. Thus, causal properties represent stored energy.

\hypertarget{structure-recombination---information-and-intelligence-are-physical.}{%
\subsubsection{Structure recombination - Information, and intelligence,
are
physical.}\label{structure-recombination---information-and-intelligence-are-physical.}}

As had been mentioned in the ``Enerstatic Loops, Networks and Nests''
section, the concept of an enerstatic loop is meant to capture the idea
that information required to reliably produce a structure is physically
embodied by such a loop. Additionally, we suggested that a structure by
itself is indeed an enerstatic loop, whose possibly varying
architectures have been abstracted away due to that structure being
deemed ``stable'', or useful to the loop in which it is embedded, only
when it is in one particular form of stasis. This paper argues that the
dynamics of enerstatic loops are to be taken as a ubiquitous consequence
of laws of physics which dictate what relationships are stable between
constituent structures at every scale. Thus, enerstatic loops are to be
understood as inter-structure relationships whose stability depends upon
(1) the interaction between the structures, and (2) the particular local
environment that the loop is in. Here, environment refers to both the
EEL and every other structure that is ``sufficiently outside'' of the
enerstatic loop in question in the sense that the structures inside the
enerstatic loop only have sparse, weak, or otherwise unstable,
relationships with structures outside the loop. The way that these
relationships form is simple. When one structure is within another
structure\textquotesingle s effective radius, a relationship is formed.
This relationship constitutes the embodiment of information about the
structures present in the relationship. The following image illustrates
what such sparse relationships could look like.

\begin{figure}[h!]
  \centering
  \includegraphics[width=6.5in,height=3.56944in]{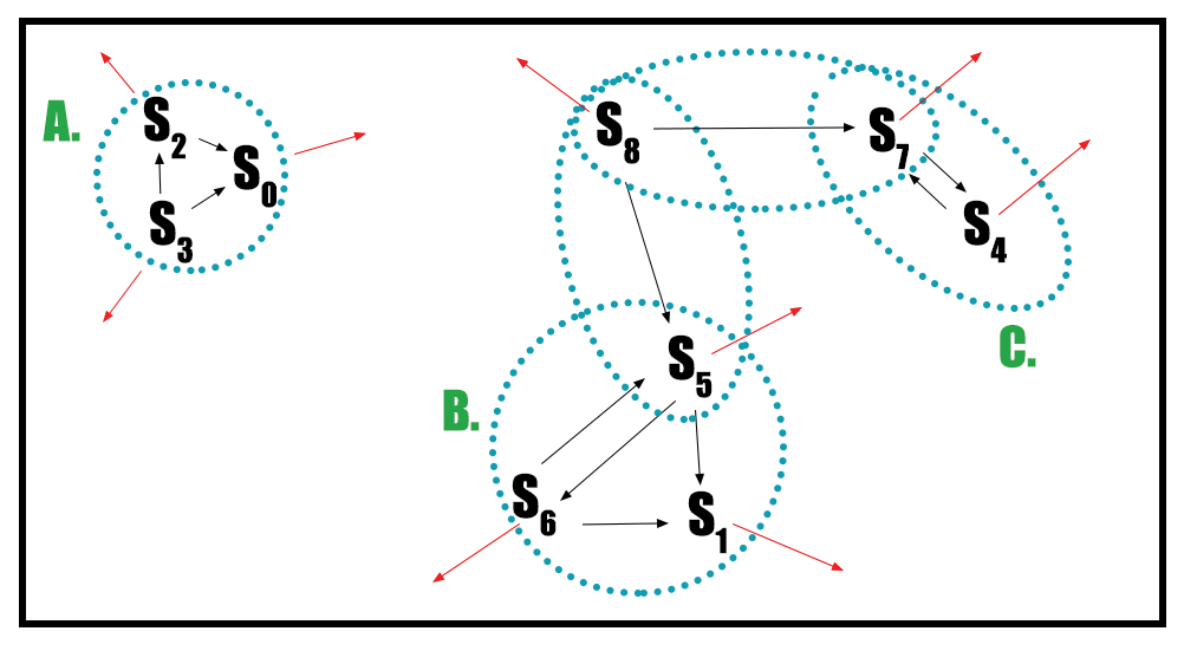}
  \label{fig:example-image}
\end{figure}

\begin{itemize}
\item
  
  Illustration of structures that have been created inside the
  ``Environmental Enerstatic Loop (EEL) and have established
  relationships with each other''. Red arrows show the impact of
  structure causal power on the environment, while black arrows indicate
  the impact of structure causal power on other structures. Dotted blue
  line indicates the enerstatic loops, which are to be understood as
  relationships between structures. Structures are in a relationship
  only when structures affect each other.
  
\item
  
  Loop A structures have no relationship with any other structure. Loop
  B \& Loop C have sparse relationships with structure
  S\textsubscript{8}, as they are affected, but don't affect
  S\textsubscript{8}.
  
\end{itemize}

As before, each enerstatic loop has an ``Enerstatic Window'' consisting
of a ``Stasis Window'', ``Action Window'' and ``Change Action
Probability Window'', however, instead of limiting loops to only be
capable of creating structures that they have already come into contact
with, we allow loops to create any structure that has been already
``invented'' by the EEL. This further breaks down the distinction
between loop and environment by treating each loop as a sort of mini EEL
in itself. Thus, what makes each enerstatic loop sufficiently different
from other loops is their differing probability distribution with
respect to what actions each loop will take given its energy level. In
this way, relationships break down when structures move too far away
from each other, when the action taken disassembles a structure, or when
energy flowing through the loop is either insufficient to sustain its
dynamics, or higher than the loop is capable of dissipating.

Defining the enerstatic window is again something that we'd like to do
automatically. For this, we take another look at our abstract syntax
tree (AST). Our abstract syntax tree allowed us to calculate the free
energy demand (FED) of any structure, but it also allows us to calculate
the free energy limit (FEL) of any structure. These define the amount of
energy a structure requires to maintain its existence, and the maximum
amount of energy a structure is capable of dissipating, respectively.
Knowing these values for every structure in an enerstatic window, as
well as their connectivity, we can create a function that allows for the
calculation of the lower and upper energy bounds of our enerstatic
window. Using such a calculation, our lower enerstatic bound is
equivalent to the smallest amount of energy required to sustain the
system, while the upper enerstatic bound is equivalent to the largest
amount of energy that can be dissipated through the network. Outside of
these bounds, the loop breaks down through the loss of structure.

\hypertarget{summarizing}{%
\subsubsection{Summarizing}\label{summarizing}}

We started off with an environmental enerstatic loop (EEL), which
consisted of a pool of ``free energy'', a large language model (LLM)
used to produce the causal structures within ``causal
niches\textquotesingle\textquotesingle, an energetic mapping used to
assign energetic costs (FEDs), and limits (FELs) to each structure, and
finally a description of the way structures make, break and adjust
relationships a la enerstatic loops. Each of these elements is flexible
when it comes to the specifics of how it is implemented. The amount of
free energy the EEL starts with, the way causal structures are
generated, the way energy costs are mapped onto logical code, and of
course, the way structures recombine and adjust their relationships, can
all be varied. Specific implementations do not reflect the Ensoul
framework. They only reflect particular ways of executing the framework.
That being said, the core idea here is to develop energetically closed
systems that generate structures which are increasingly interesting,
intelligent, and ``alive'' as generally defined in the ``Open-Ended
Evolution and Assembly Theory'' section.

Quantum computing demonstrates that it may not be practical, or even
desirable, to physically create perfectly energetically closed systems,
however, it might be that approximations to such systems would serve as
powerful ways to both more easily generate intelligent technology in the
physical world, and serve as models for understanding the evolution, and
dynamics, of thermodynamically open systems more deeply. Before diving
into creating the physical manifestations of such devices, it is
important to first demonstrate their capabilities in the computer, and
use such model systems to inform the creation of them.

What we've effectively done in this paper is to make code physical by
creating an underlying ``energetic physics'' on top of which our code is
built. In other words, ``structures'' are code that has been manifested
into an energetically constrained virtual realm. This is in alignment
with the idea that everything in the physical world can be said to be
``computing'', and is therefore describable via computations possible by
a universal Turing machine. This has been called pancomputationalism
\textbf{\cite{59}}. While it is not clear that everything in the physical
world is computable; it is also not clear that we have any other choice
but to use mathematical and computational tools to understand it.

So far, the discussion has been about what we can do inside of a
computer in order to create SOULS. These simulations are important for
the development of software based technology, in its own right, however,
simulations of energetically closed systems using conventional
Von-Neumann architecture IOS severely underutilize the true
computational capacity of the materials being used. This underutilized
computational capacity, is what we referred to in ``IOS Illusions:
Motivating Modeling With Enerstatic Loops'' as ``highly variable
behavior'', or ``noise''. This property of materials that engineers have
painstakingly avoided can actually function as useful computational
processes if one takes the right approach. Systems that embrace noise as
a computational resource have been called ``Thermodynamic Computers''
(TCs). These computers are SOULS in physical reality and truly bring us
closer to the ``Ultra Low power'' part of the SOULS definition.

\hypertarget{thermodynamic-computing-enerstatic-networks}{%
\section{Thermodynamic Computing \& Enerstatic
Networks}\label{thermodynamic-computing-enerstatic-networks}}

Thermodynamic computers have been envisioned as an ``engineered,
multi-scale, complex system that, when exposed to external potentials
(inputs), spontaneously transitions between states in the short term
while refining its organization in the long term, both as part of an
inherent, adaptive process driven by thermodynamics ``
\textbf{\cite{40}}. This is analogous to what we've described as the
behavior of enerstatic networks, but in physical form. This paper argues
that the collection and study of manifestations of this general concept
will prove to be important for understanding the problem of embedded
agency. In this section, we will briefly touch on thermodynamic
computers, the relationship between thermodynamic computing and
enerstatic networks, and finally, embedded agency.

Thermodynamics is a branch of physics that deals with the relationships
between the various different forms of energy such as heat and work. It
was originally developed in order to optimize the amount of energy
extracted from steam engines \textbf{\cite{41}}. This application of
thermodynamics focuses on states of ``equilibrium'' in which the state
of a system is homogenous and macroscopically unchanging, such as a pool
of still water. Indeed, one of the conclusions drawn from thermodynamics
is that the entire universe is evolving towards such an equilibrium
state in the so-called ``heat death of the universe''. In spite of this
conclusion, many physicists have come to believe that thermodynamics is
also chiefly responsible for driving the organization of life as we know
it \textbf{\cite{60, 61}}. This has spurred the development of
``Non-equilibrium Thermodynamics'', which is thought to be the type of
thermodynamics that life, and many other complex systems that are far
from equilibrium, seem to employ. In particular, work on
fluctuation-dissipation theorems and information thermodynamics have
been continuing to generate insights into understanding life and the
development of thermodynamic computers.

One of the issues in thermodynamic computing is the lack of a unifying
general model system that can properly express thermodynamic computing.
While there are existing model systems such as Hopfield Networks,
Boltzmann Machines, thermodynamic neural computation models, and
Thermodynamic Neural Networks; These models have been regarded as only
``a small set of interesting and suggestive devices and architectures
\textbf{\cite{41}}''. Although there remains much work to be done with
respect to this area of research, this paper suggests that enerstatic
networks may be a general enough model to serve as such a unifying
general model system. This kind of system is critical for the evaluation
of core theoretical concepts, like fluctuation-dissipation theorems, and
information thermodynamics, the demonstration of problems that are best
solved using such approaches, and as a helpful reference for the
development of thermodynamic computers.

The reasoning behind speculation that enerstatic networks might serve as
such a model is based on the fact that their components -- enerstatic
loops -- are ultimately extremely simplified abstractions of
thermodynamically open systems whose level of detail can be
straightforwardly scaled up through the use of empirically measured
values and integration of empirically derived mathematical equations.

Given the existence of SOULS, whether they are virtual, physical,
engineered, or evolved, and the need to control, or take advantage of
said SOULS, one needs a proper framework, or way of thinking, with which
to do so. This brings us back to TAME -- A Technological Approach to
Mind Everywhere -- which provides many such ideas and beliefs in order
to accomplish just that.

\hypertarget{taming-souls}{%
\section{\texorpdfstring{T\textbf{aming
SOULS}}{Taming SOULS}}\label{taming-souls}}

While it is trivially true that the goal of any ``thing'' is to exist,
it is apparent that ``things'' also seem to have ``higher level'' goals
that very much seem to have little to nothing to do with existence in
particular. For example, a person who chooses to summit Mt. Everest or
to run in a 135 mile race through death valley in the scorching hot sun
doesn't quite seem to be taking the most optimal path towards a
continued existence.

\hypertarget{deceptive-higher-level-goals}{%
\subsection{Deceptive higher level
goals}\label{deceptive-higher-level-goals}}

This paper argues that the pursuit of existence can be deceptive in a
way that is perhaps analogous to the way the path to a particular goal
can be deceptive as we discussed with the training of the biped. That is
to say, the path to existence appears to be divergent rather than
convergent. The invariant multi-scale TAME approach suggests an
interesting way of describing this seeming inconsistency.

First of all, any ``thing'' is made up of parts whose goal is to exist.
For many things, particularly those we consider ``more alive'', those
parts are not densely connected to each other. In other words, each part
is not directly causally connected to every other part. Considering that
the coherence of a ``self'' is determined by the degree to which parts
share stress, certain parts can be partitioned as belonging to different
``things'', or in other words, to some degree, constituting separate
''selves''. A relatable example is the fact that biological organisms
are made up of cells (parts), which cluster together in such a way as to
be called different tissues (things), which then differentially cluster
to be called organs, and so on. As previously discussed, when resources
are limited, achieving the goal of existence necessarily induces
resource limitations for other ``things'', since there is less free
energy to utilize. So, the thing about not directly sharing stress with
other parts, is that it now becomes possible for parts of any ``thing''
to compete with each other over resources despite ultimately belonging
to the same ``thing''.

However, the possibility of competition does not imply competition must
be happening. In other words, despite having the ability to strictly
compete and perform actions that would be detrimental to the collective
as a whole, persistent collectives of cells achieve consensus through a
combination of cooperation and competition, which we will refer to here
as ``coopetition''. The consensus that arises from coopetitive dynamics
are rather nicely illustrated by financial markets, and companies, in
which individuals, and populations of individuals, only acting according
to their own local interests, are able to achieve stable market prices
and can ensure that all involved balance their books. Coopetition
operates at all scales of collective intelligences, from organelles, to
cells, to tissues, to organs and so on. According to Dr. Stephen
Grossberg's work in such coopetitive systems, there is a trade off
between how free parts are to make certain decisions, and the size of
the global decision that can be made. This restriction can result in the
existence of multiple choices in a network, in which a novel choice can
be preferred, and in oscillatory, or ``jumps'', between different
decisions, or parts, that are currently ``winning'' the competition
\textbf{\cite{41, 42, 43, 44}}. Both of these phenomena contribute to interesting, complex
adaptive behavior as the collective in question takes actions which in
favorable cases allow the collective to survive. Dr. Grossberg proved a
theorem about this dynamic that allows for the design of such systems
called the ``Adaptation Level System theorem''. As an incredibly
simplistic example, one might understand this in the following way: For
our climber, reaching the summit of Mt. Everest is bad for some
populations of cells in the collective, but great for the survival of
other populations. The populations of cells that benefit from this
behavior are ``winning'' the competition, but this is balanced by yet
other populations ``winning'' the competition once the mountain has been
descended thus all parties involved balance their books.

Now we have an idea behind what might in many cases be the cause of
seemingly existentially-irrelevant behavior of ``things''. Essentially,
it is dependent on the coopetition between parts within a ``thing'',
which can be called ``parts'' as a function of the degree to which their
constituent pieces share stress. The task is then to figure out how one
can identify the goals, or purpose of the decisions, made by the
collective. Such an endeavor enables us to better characterize, engineer
and work with such collectives.

\hypertarget{taming-souls-and-cognitive-capacity-tests-ccts}{%
\subsection{Taming SOULS and Cognitive Capacity Tests
(CCTs)}\label{taming-souls-and-cognitive-capacity-tests-ccts}}

Perceptual Control Theory (PCT) is a version of control theory that is
dedicated to understanding naturally occurring control systems, or
SOULS. PCT asserts that:

``A natural control system can be organized only around the effects that
its actions (or independent events) have on its inputs (broadly
defined), for its inputs contain all consequences of its actions that
can conceivably matter to the control system \textbf{\cite{48}}''.

Powers well describes this concept in just a few words saying that
``behavior is the control of perception (inputs)''. In other words,
actions are taken in service of ensuring that the ``thing'' perceives
what it ``wants'' to perceive. In PCT, this desired perception is called
the ``reference signal'', and critically, is thought of as being
internally generated as opposed to externally generated as it is in most
artificial control systems. As an example of this externality, the value
of the temperature that a thermostat is trying to regulate, depends on
the reference signal given to it by its user and typically is not
modifiable by the thermostat itself. For a natural control system the
reference signal is always modifiable by the ``thing'' itself. Consider
again the non-analytical solution to the outfielder problem known as
``Optical Acceleration Cancellation''. The natural control system, in
this case the outfielder, presumably comes to choose the variable that
they are controlling. That same control system can work towards
controlling a number of other variables all towards solutions of
different tasks making its position on the cognitive continuum much
farther to the right than a simple thermostat.

\hypertarget{test-for-controlled-variable-tcv}{%
\subsubsection{Test for Controlled Variable
(TCV)}\label{test-for-controlled-variable-tcv}}

The protocol for inferring the controlled variable of any control system
is the same whether it is natural, or artificial. Under the PCT, this
protocol is called the ``Test for Controlled Variable (TCV)''. TCV
allows one to identify both the type of system that one is looking at
(IOS vs SOULS), and, if present, the variable(s) that the SOULS is
controlling. The TCV process is an iterative empirical process through
which a hypothesis about what variable is being controlled is tested.
The hypothesis takes the following form: ``If variable X is the variable
under control, then variable X should experience less than usual effect
upon being disturbed''. Temperature under the control of a thermostat
provides a straightforward example. Provided your AC was on, opening a
window on a chilly day would slowly cause a decrease in temperature,
which would soon be countered by the thermostat's activation of the
heating mechanism. This is not what would be expected if the thermostat
was not controlling temperature as a variable and constitutes the
temperature experiencing a ``less than usual effect upon being
disturbed''. A non-trivial application of TCV was in the identification
of ``control of optical velocity (COV)'' over ``optical acceleration
cancellation (OAC)'' as the better explanation for how organisms solve
the outfielder problem \textbf{\cite{40}}. These two variables, velocity
and acceleration are closely related, as such, it is difficult to
disturb one without disturbing the other. In order to help sway
interpretations in a single direction, two model control systems were
made that each implement control of only one of these variables. The OAC
model accounted for 75\% of the variance in the path of an organism
intercepting an object, while the COV model accounted for 93\% of the
variance. This suggests that optical velocity, rather than optical
acceleration, is the variable actually being controlled during object
interception. TCV rules out the possibility that the system we are
studying is an IOS by confirming that no variable is under the control
of that system's behavior. In the case of something like our ``odd
thermostat'', which is a thermostat that regulates temperature with its
output only 50\% of the time, the application of statistical methods
along with TCV can allow us to gauge where on the IOS vs SOULS continuum
our system lies.

While TCV allows us to figure out what variable the system is
controlling during some particular task, it does not tell us the entire
set of variables a control system is capable of controlling, nor does it
tell us how we might modify the value of reference signal of the systems
controlled variable, or how we might induce the system to switch between
different controlled variables. Such inquisition might lead us to coming
up with a ``Test for Variable Controllability (TVC)'' and a ``Test for
Control Switches (TCS)'' respectively. We'll now briefly discuss how
these could work.

\hypertarget{test-for-variable-controllability-tvc}{%
\subsubsection{Test for Variable Controllability
(TVC)}\label{test-for-variable-controllability-tvc}}

Test for Variable Controllability (TVC) attempts to answer the question:
``What variables that I am capable of measuring are this system capable
of sensing?''. This framing makes it clear that the observer is chiefly
responsible for properly inferring the system's cognitive capacity. Like
TCV, TVC starts with a hypothesis about some set of measurable variables
lending a credence to each one. The testing amounts to sensor
identification and subsequent inference about what sensory partitions
the control system might make. In other words, identifying the location
and type of sensor, as well as what these sensors could amount to
sensing as a collective (i.e. light detection (small scale), vs
perception of an entire object (large scale)). A simple example is
inferring that based on having eyes a creature might be able to detect
individual objects. Although non-trivial, the task is important to
mapping out the spatial dimension of a system\textquotesingle s
``cognitive light cone'', which is a system\textquotesingle s
spatiotemporal computational capacity towards attaining a particular
goal. The temporal dimension is intimately related to the spatial
dimension. In TVC we ultimately care about what type of measurable
variable the system expends energy towards controlling as well as about
how far disturbances can be in space and time and still evoke a change
in the system's actions. In short, the TVC should work towards mapping a
system\textquotesingle s cognitive light cone by starting from the
theoretical maximum and conducting experimental tests to cut down on
possibilities.

\hypertarget{test-for-control-switch-tcs}{%
\subsubsection{Test for Control Switch
(TCS)}\label{test-for-control-switch-tcs}}

Test for Control Switch (TCS) is a test that is meant to identify the
method that one should use to modify the value, and/or identity, of the
variable that the system is controlling. TCS is related to the ``axis of
persuadability'' concept presented by TAME. Recall that the position of
a system on the axis of persuadability represents the level of
complexity at which intervention is required to get a system to do what
you'd like it to do. The farther to the right the system is on the axis
of persuadability, the less energy one has to expend. In other words, a
good control switch is one that ultimately allows one to expend as
little energy as possible. As such, TCS should test for switches at
different levels of cognitive capacity starting from the highest
potential cognition whose cognitive light cone was inferred via the TVC.
In short, TCS would ask:

\begin{enumerate}
\def\labelenumi{\arabic{enumi}.}
\item
  
  Can I communicate with the system with some language through some
  medium?
  
\item
  
  Can I identify rewards for this system which to use for training
  purposes?
  
\item
  
  Are their setpoints which I could modify using some physical means?
  
\item
  
  Would I really have to reduce it into pieces and refashion it together
  from scratch to get it to do what I want?
  
\end{enumerate}

The emphasis here is to avoid underestimating the intelligence of a
system. One excellent example demonstrating the power of a control
switch is found in the demonstration of morphological control through
modification of a bioelectrically embodied setpoint \textbf{\cite{62}}.
The lab of Dr. Michael Levin, and their collaborators, established the
bioelectrical state of cells as instructive to their collective target
morphology through a few key examples. First, it helps to know that
cells are typically more negatively charged inside the cell than outside
the cell. The charge, or voltage, inside a cell can be measured and
visualized. This allows us to see the impact of certain drugs, or other
interventions on the collective bioelectrical state of some cells. The
first of Levin's experiments that we will mention is in flatworms.
Flatworms are complex organisms that exhibit bilateral symmetry and have
a nervous system. They also have spectacular regenerative abilities.
Flatworms can be cut in two, which causes each half to develop into
perfectly good flatworms. A piece of flatworm as small as 1/279th the
size of the animal has been demonstrated to regrow into a fully viable
adult flatworm \textbf{\cite{47}}. Through visualizing the bioelectrical
state of the cells one can actually induce a severed planaria to have
two heads rather than one by manipulating the cells of the ``should be''
tail. Two tailed flatworms have been made in this way, flatworms with
heads from other flatworm species, and even flatworms with heads that
are presumably non-existent in flatworm nature as we've known it thus
far. Importantly, once these bioelectrical manipulations are made, the
flatworm remembers the state upon being severed once again. In other
words, once severed, a two headed flatworm begets two, two headed
flatworms. This is despite the genome not being modified in any way.
This represents a powerful control switch for biological organisms. In
the same vein was in the case of tadpoles which can be bioelectrically
manipulated in order to have an eye on their tail. Not only is an eye on
the tadpole's tail with no genetic modification, the cells of the
tadpole actually organized themselves towards making the eye functional
from a relatively simple local signal \textbf{\cite{63}}. Tadpoles
original eyes were blinded to demonstrate that it could still navigate
with the tail-eye \textbf{\cite{64}}. Further evidence of bioelectrical
cell state as a control switch was in the creation of so-called
Xenobots. Xenobots were created by Levin in collaboration with Dr. Josh
Bongard. Xenobots are simply what one gets when the skin cells of a frog
are liberated from the electrical signals they normally get from the
rest of the body during development. Once liberated, these cells
coagulate to form a proto-organism that reproduces itself by physically
moving other cells around it into similar proto-organisms. These
phenomena make it clear that bioelectrical states of cells act as a
setpoint that controls their behavior. Indeed, Levin demonstrated that
certain cancers can be effectively cured by forcing cancer cells to
reintegrate electrically with the collective \textbf{\cite{65}}.

Identifying a system\textquotesingle s cognitive light cone and control
switches are both big tasks, but the reward certainly justifies the work
involved. Controlling a system by rewiring its hardware should only be
resorted to when absolutely necessary. The competence of naturally
occurring control systems should make it clear that the intelligence of
the parts that make up natural control systems should be properly
assessed in order to determine the most efficient way to understand and
leverage its behavior towards a particular goal. Importantly, these
tests all have usages towards the understanding and development of
thermodynamic computers. Making all of these tests more rigorous and
formalizable is an aim of TAME and a major motivation for this paper. A
model that can capture the self-organizing dynamics of thermodynamically
driven complex adaptive systems would serve as the ideal test bed for
theoretical calculations in the development of empirical
tests/statistics and in the development of more biologically, or rather,
thermodynamically, inspired technological advancements. We now recap the
desired perception(s) driving all of this behavior and discuss future
directions.

\hypertarget{discussion-research-directions-speculations-and-philosophy}{%
\section{Discussion: Research Directions, Speculations and
Philosophy}\label{discussion-research-directions-speculations-and-philosophy}}

\hypertarget{summarizing-ensoul}{%
\subsection{Summarizing Ensoul}\label{summarizing-ensoul}}

``Given that we see neurons that exist over the lifetime of an organism,
what must it be really good at doing in order to do so?'' -- This paper
suggested that that thing was ``managing its energy level'', in which
case, analysis of its behavior assuming any other purpose would
frustrate such analysis. However, information processing certainly seems
to be just what neurons do. Thus, is it possible to build systems in
which information processing, or ``work'', is done as a consequence only
of a ``thing'' managing its energy level? Might this concept relate more
generally to what we observe in physical reality? -- This paper answered
affirmatively to both of these questions. The generalization of this
concept led to the thermodynamically inspired abstraction, the
enerstatic loop. The core driving force being that, any ``thing'' which
exists, and persists in its existence, must be participating in
something that allows for that persistent existence. In other words, it
must have causal power, and that causal power must either be embedded in
the self-evidencing behavior of some other ``thing'', or must itself
exhibit some degree of self-evidencing behavior. This scheme implies
that any persistent thing is well understood as a control system in
which negative feedback maintains what would otherwise be broken down
and that this dynamic is intrinsically generated by the parts which make
up that system.

Following such a scheme, this paper leaned into pancomputationalism, the
idea that such causal power can always be written down formally and is
computable. However, actually writing such code by hand is non-trivial
due to the simultaneous considerations of energy and information
processing. Additionally, the findings in open-ended evolution make it
clear that the path to a desired outcome isn't always best found with
optimization (convergent) techniques. Fortunately, enerstatic loops are
well suited to open-ended evolution due to their existential imperative
and capacity for information processing. Mapping energy costs to code (A
process one might call ``deabstracting'', or ``physicalizing''), the use
of large language models, the power of assembly theory as basis for
complexity measures in evolutionary systems, and the noted information
symmetry posited by the free energy principle makes the exploration of
``energetically closed systems'' a promising avenue for the
understanding, and generation of, models of thermodynamically open
systems based on enerstatic loops. Given the abstract
substrate-independent nature of such energetically closed systems, and
their generated energetically open systems, the term ``Self Organizing
intelligent Ultra Low Power System (SOULS)'' seemed a fitting
description for both such entities.

The understanding of ``things'' such as SOULS illustrate the need for an
existentially-based multi-scale perspective on reality, a perspective
which focuses on the invariant concepts that can be consistently applied
to every ``thing'' at every scale. Technological Approach to Mind
Everywhere (TAME) serves as an important framework for taking such a
perspective. As a result, the concepts of ``intelligence'',
``cognition'', ``goals'', ``stress'', ``self'', ``life'', and
``feedback'' are defined on a continuum, rather than with discrete
categories. We ask not if the ``thing'' has these properties, but of
``what kind'', and of ``how much'' \textbf{\cite{5}}. This is critical
for developing an understanding of complex systems, one that
specifically allows us to understand how these concepts scale up, and
what the macroscopic result of such scaling looks like.

It is clear that the influence of technological advancements on humanity
have been increasing exponentially in their rate of development with no
apparent end in sight. The results of these advancements have often been
unpredictable with both positive and negative consequences. As such, the
development of the technology described in this paper is meant to serve
two purposes: (1) The generation of practical technologies that can be
put to use and, (2) to serve as a test bed through which negative
consequences can be better understood and mitigated. Understanding the
consequences of such technology is a non-trivial task, but it is
critical that we be able not just to build it, but to control it. Lack
of understanding of the consequences of our current technologies have
resulted in things like loss of privacy, addiction to social media and
unprecedented political influence. At the same time, there are strong
arguments that the overall quality of life, including for those in
impoverished conditions, has significantly improved as a result of
technological progress and increased global interconnectedness. This
makes the development of a framework like TAME alongside the development
of SOULS both an existential, and utopic, imperative. The faster the
development of our understanding, the better chances we have of
intentionally shaping the consequences of our technological
advancements. This leads us to considering important future research
directions.

\hypertarget{future-directions}{%
\subsection{Future directions}\label{future-directions}}

Briefly, future directions include:

\begin{enumerate}
\def\labelenumi{\arabic{enumi}.}
\item
  
  Using conventional von-Neumann computers to implement enerstatic
  programs, perhaps via FPGAs, or other flexible classical computing
  approaches.
  
\item
  
  The development of specialized versions of enerstatic programs aid as
  a test bed in the development of thermodynamic computers through
  advancement of theoretical/mathematical frameworks.
  
\item
  
  The development of specialized versions of enerstatic programs to
  assist in the understanding of biological systems, an example of which
  was the enerstatic neural network described in this paper.
  
\item
  
  The development of rigorous measures of intelligence, interestingness,
  and ``aliveness'' (complexity) in energetically closed enerstatic
  programs such that different programs can be compared and autonomously
  analyzed.
  
\item
  
  The development of TAME, and more specifically, the development of
  rigorous cognitive capacity tests (CCTs), (i.e.. existential measures
  of intelligence/cognition, test for controlled variable (TCV), test
  for variable controllability (TVC), and test for control switch (TCS).
  
\item
  
  Finally, if deemed necessary, the creation of alternative methods than
  the use of Large Language Models for the generation of structures in
  energetically closed systems.
  
\end{enumerate}

\hypertarget{speculations-and-philosophy}{%
\subsection{Speculations and
philosophy}\label{speculations-and-philosophy}}

The contents of this paper were written in an attempt to put together
ideas that may be key to advancing the understanding and development of
science and technology. To that end, this paper embraced substance
monism, process philosophy, pancomputationalism, and an extended version
of the strong life-mind continuity thesis. Philosophy is essential
because the way that we think of the world influences the way that we
act in that world, which importantly, includes how we conduct scientific
research. As such, we now step through some of the philosophy motivating
the development of Ensoul.

Substance monism suggests that everything is made up of ``one thing",
which seems to then motivate the question: ``Why do we see distinct
`things' if everything is made up of one thing?'' and ``How do these
things come into being?''. While this paper provides no answers beyond
``it is just so'', it does suggest physically and logically grounded
guidelines as to how to simulate the behavior of such ``things'' as well
as how to go about evolving systems with such ``things''. The following
argument suggested a way about how to go on:

\begin{enumerate}
\def\labelenumi{\arabic{enumi}.}
\item
  
  Physics has told us that E=mc\textsuperscript{2}, in other words, that
  every ``thing'', is made up of energy.
  
\item
  
  The way that we understand the behavior of ``things'' is through cause
  and effect.
  
\item
  
  A ``thing'' cannot be said to exist, if it has no effect. In other
  words, to exist is to be the cause of an effect (eleatic principle).
  
\item
  
  Similarly, a ``thing'' cannot be said to be persisting its existence
  if there is no cause(s) for its existence. In other words, to persist
  is to be the effect of a cause (Persistence principle).
  
\item
  
  Therefore, we are to model ``persistent things'' as having persistent
  causes, of which they themselves may be the cause. In other words, to
  model their existential and self-evidencing causal powers.
  
\end{enumerate}

This leads to a continuous vision of ``things'' that have various
degrees of what one might call ``existential feedback''. This is what
happens when a ``things'' effect is part of its cause. Indeed, if this
is not the case, then it must be the case that something else causes it,
which itself must be caused by itself, or something else. Therefore, we
form the hypothesis that things which cause their own existence have a
more prevalent/sustained existence. These ``things'' are well understood
as negative feedback systems, or control systems. The interesting thing
about the nature of these natural control systems, is that they are
regulating their own particular dynamics, or causal power, rather than
any one particular measurable variable. This view is well in line with
process philosophy which suggests no ``thing'' is static, but is always
changing. This paper uses this necessary dynamic of ``changing'' in
combination with the eleatic principle and principle of causality to
motivate existential circular causality, or existential feedback loops,
as proper abstractions for such systems. In doing so, it leans into
pancomputationalism and the continuity thesis proper.

Pancomputationalism, like process theory, suggests that the physical
world can be understood as a process, however, it specifically suggests
that this process is a computational one. That is, a cause and effect
process that can be simulated on a computer. The reason this paper leans
into this notion is two fold: (1) If we are to understand the nature of
reality we must use the language of cause and effect as its automation
in computational devices has helped extend our cognition in ways we
could have never thought possible, and (2) This sort of approach might
provide strong evidence that pancomputationalism cannot be sufficient
through the demonstration that properties which we typically describe as
``emergent'' truly cannot be described as being computationally
attainable by the fundamental constituents of that phenomenon.

In leaning towards the idea that these properties can indeed be found as
computationally attainable by the dynamics of the fundamental ``stuff''
that makes up the universe, this paper embraces the strong life-mind
continuity thesis \textbf{\cite{67}}. The strong life-mind continuity
thesis suggests that the origin of life is synonymous with the origin of
mind. In other words, once you have life, you have a mind. This paper
attempts to operationalize familiar properties such as intelligence and
cognition as dependent on a ``things'' self-evidencing causal power. It
is on these grounds that the efforts encouraged by this paper rest.

The challenge at hand is to use the abstraction of an existential
feedback loop as a tool to understand how such operationally defined
properties scale up into the forms that we are familiar with, as well as
discover how the dynamics of such loops can be leveraged in the creation
of new technologies and understanding of life as it might be. To this
end, embracing evolutionary approaches along with traditional approaches
to the creation of enerstatic loops, networks and nests in virtual and
physical spaces may prove to be a fruitful area of exploration.

A final word, one should be careful not to mistake the map for the
territory. Philosophy, theories and models of reality are not equivalent
to reality despite their predictive power or believability. The
situation that we find ourselves in is one in which we seemingly cannot
truly know what is behind the causes of our senses \textbf{\cite{49}}.
Philosophy, theories and models are ultimately just stories we tell
ourselves about those possible hidden causes. SOULS is a term that just
so happens to fit well for any "thing" that persists its existence while
abiding by the stories asserted by process philosophy and substance
monism. As such, we can ask whether taken metaphysically, the notion of
SOULS can have application in giving the colloquial word ``soul'' a more
rigorous definition. Philosophically speaking, Ensoul would seem to
suggest that a soul is any causal force that is persistent over time.
Thus, approaches like Ensoul, which use existential feedback loops may
be considered a sort of implementable philosophical model, one in which
this philosophical story might be better understood. Stories represent
our subjective beliefs about what is happening in reality and as such
the stories that we tell ourselves guide what we do in both life and
science. Thus, while it is important to acknowledge that these stories
are not equivalent to reality, these stories, should we believe them,
are equally important, considering the power that these stories have
over the sorts of actions that we take in the world. Interestingly, the
word ``psychology'' was originally defined as ``the study of souls''
before becoming more commonly recognized as the ``study of minds''.
Indeed, the word ``psychology'' is made from the combination of the
Greek words ``psychē `` (meaning ``breath, principle of life, life,
soul''), and ''logia'' ( meaning ``speech, word, reason''). One of the
aims of Ensoul is to take seriously the idea that some of the properties
one studies in psychology, like intelligence and cognition, are not just
present in the human mind, but in all ``things'', or SOULS. As such,
methods and ideas from psychology such as PCT and reinforcement learning
can be useful in understanding the nature of any ``thing'' independent
of what that ``thing'' is made up of. This might suggest an extension,
or reinterpretation, of psychology that one might call
``neopsychology'', or the ``new study of the SOULS'', where what we are
primarily interested in are these measurable substrate-invariant
operationally defined properties of any ``thing'' (i.e. intelligence,
cognition, stress level, etc) and how they scale up, or in what cases,
if any, certain properties can be said to "emerge". Since this paper
primarily takes a modeling, or construction approach to understanding
SOULS, the framework has been named ``Ensoul'' which is a verb meaning
``to endow with a soul''.

\newpage

{\centering
  \bibliography{citations} 
\par}

\hypertarget{section-1}{%
\section{}\label{section-1}}

\hypertarget{section-2}{%
\section{}\label{section-2}}

\hypertarget{section-3}{%
\section{}\label{section-3}}

\hypertarget{section-4}{%
\section{}\label{section-4}}

\newpage

{\centering
  \hypertarget{glossary}{%
  \section{Glossary}\label{glossary}}
\par}

\begin{itemize}
\item
  
  \textbf{Assembly Theory} - A physical theory of life which allows one
  to categorize and detect life through measurement and application of a
  recursive algorithm. The theory places different life forms on a
  spectrum of complexity.
  
\item
  
  \textbf{Axis of persuadability} - the notion that every ``thing'' lies
  on a continuum with respect to the amount of effort it takes to
  control that thing. The amount of effort required is inversely
  proportional to the complexity/size of the ``things'' cognitive light
  cone.
  
\item
  
  \textbf{Cognition} - Operationally defined in this paper as the degree
  to which a ``thing'' can ``sense'', ``believe'' and ``act''. In other
  words, the complexity associated
  
\item
  
  \textbf{Cognitive Capacity Tests} - Describes a suite of tests used to
  determine the cognitive light cone of any ``thing''. Some tests
  include: Test for controlled variable (TCV), test for variable
  controllability (TVC), and test for control switch (TCS).
  
\item
  
  \textbf{Cognitive Light Cone} - describes how far in space and time an
  event can be and still have an affect on the computations being
  enacted by some system.
  
\item
  
  \textbf{Enerstatic Loop} - Short for ``Energy Homeostatic Loop'',
  which is an abstraction of a control system that learns to regulate
  its own ``energy level'', or particular dynamics, through the
  modification of its own structure. It maps onto the idea of an
  adaptive thermodynamically open system, or adaptive dissipative
  structure. Enerstatic loops represent ``relationships'' between
  ``structures''
  
\item
  
  \textbf{Ensoul} - Acronym for ``Enerstatic Network approach to Self
  Organizing intelligent Ultra Low Power Systems'', a framework that
  this paper proposes in order to better understand the nature of
  reality.
  
\item
  
  \textbf{Environmental Enerstatic Loop (EEL)} - Short for
  ``Environmental Enerstatic Loop'', which is an abstraction utilized to
  implement energetically closed systems. The EEL is responsible for
  creating structures and enerstatic loops.
  
\item
  
  \textbf{Feedback} - The degree to which a thing's effect, or
  ``output'', affects its cause, or ``input''
  
\item
  
  \textbf{Framework} - A framework is a set of rules, ideas, or beliefs
  that you employ in order to solve problems or make decisions.
  
\item
  
  \textbf{Free Energy Principle} - A mathematical formalization of the
  properties that any self-organizing system must have.
  
\item
  
  \textbf{Goal} - The desired result of a thing's behavior (i.e. causal
  power).
  
\item
  
  \textbf{Intelligence} - Operationally defined in this paper as the
  adaptive capacity of a ``thing'' to meet its ``goal'' of existence.
  
\item
  
  \textbf{Life-Mind Continuity Thesis} - The philosophical idea that
  life and mind emerged simultaneously and have since then manifested
  themselves in increasingly complex ways.
  
\item
  
  \textbf{Neopsychologist} - ``the new study of SOULS'' - potential term
  for one who uses the framework and approaches written in this paper.
  Specifically, one who is primarily interested in studying measurable
  substrate-invariant operationally defined properties of any ``thing''
  (i.e. intelligence, cognition, stress level, etc).
  
\item
  
  \textbf{Open ended evolution} - The process through which new
  ``things\textquotesingle\textquotesingle{} are generated for a
  seemingly indefinite period of time. This process is believed to be
  embodied by our universe and is well described by assembly theory.
  
\item
  
  \textbf{Pancomputationalism -} The metaphysical idea that every
  ``thing'' that exists can be understood as ``computing'' something.
  
\item
  
  \textbf{Self} - A collection of parts that share degrees of stress
  
\item
  
  \textbf{Self-Evidencing} - The verb form of ``self-evident''; an
  explanation is self-evidencing if the event which it explains provides
  key evidence for the explanation. What any ``thing'' does provides key
  evidence for its own existence via an existential feedback loop.
  
\item
  
  \textbf{SOULS} - Acronym for ``Self Organizing intelligent Ultra Low
  Power System(s)'' meant to capture the multi-scale substrate
  independent nature of ``things'' that Ensoul focuses on understanding
  and utilizing to create novel technologies.
  
\item
  
  \textbf{Stress} - Any effect that pushes a ``thing'' away from its
  ``preferred'', or ``healthy'', state.
  
\item
  
  \textbf{Substance Monism -} The metaphysical idea that every ``thing''
  is made up of the same sort of ``stuff''
  
\item
  
  \textbf{IOS} - Acronym for ``Input Output System(s)'' meant to capture
  SOULS whose ``fluctuations'', or, goal-directed behavior, have been
  squelched.
  
\item
  
  \textbf{TAME} - Acronym for ``Technological Approach to Mind
  Everywhere'' a framework developed by Michael Levin to understand
  diverse bodies and minds.
  
\item
  
  \textbf{Teleology} - Purposeful, or goal directed, behavior.
  
\item
  
  \textbf{Teleonomy} - Behavior that is not truly goal directed, but
  looks ``as if'' it is.
  
\item
  
  \textbf{von-Neumann architecture} - A computer architecture based on a
  1945 description by John von Neumann. This is the architecture that
  most computers today are based on. It has evolved to mean any
  ``stored-program computer'' in which an instruction fetch and a data
  operation cannot occur at the same time.
  
\end{itemize}

\end{document}